\documentclass[journal]{IEEEtran}
\usepackage{amsmath}
\usepackage{cases}
\usepackage{graphicx}
\usepackage{float}
\usepackage{subfigure}
\usepackage{makecell}
\usepackage{textcomp}
\usepackage{array} 
\usepackage{longtable}
\usepackage{booktabs}
\usepackage{threeparttable}
\usepackage{makecell}
\usepackage{stfloats}
\usepackage{multirow}
\usepackage{cite}
\usepackage{color}

\begin{document}
	
	\title{A Synapse-Threshold Synergistic Learning Approach for Spiking Neural Networks}
	
	\author{Hongze Sun, Wuque Cai, Baoxin Yang, Yan Cui, Yang Xia, Dezhong Yao, \emph{Senior Member, IEEE}, \\and Daqing Guo  
		\thanks{This work was supported in part by the STI 2030--Major Project under Grant 2022ZD0208500 and in part by the National Natural Science Foundation of China under Grant 31771149, Grant 61933003, and Grant 82072011. (Corresponding authors: Dezhong Yao; Daqing Guo).
			
			Hongze Sun, Wuque Cai, Baoxin Yang, Yang Xia, Daqing Guo are with the Clinical Hospital of Chengdu Brain Science Institute, MOE Key Lab for NeuroInformation, School of Life Science and Technology, University of Electronic Science and Technology of China, Chengdu 611731, China (e-mail: dqguo@uestc.edu.cn). 
			
			Yan Cui is with Department of Neurosurgery, Sichuan Provincial People’s Hospital, University of Electronic Science and Technology of China,
			Chengdu 610072, China
			
			Dezhong Yao is with the Clinical Hospital of Chengdu Brain Science Institute, MOE Key Lab for NeuroInformation, School of Life Science and Technology, University of Electronic Science and Technology of China, Chengdu 611731, China, also with the Research Unit of NeuroInformation (2019RU035), Chinese Academy of Medical Sciences, Chengdu 611731, China, and also with the School of Electrical Engineering, Zhengzhou University, Zhengzhou 450001, China (e-mail: dyao@uestc.edu.cn).}
	}
	\markboth{}%
	{}

	\maketitle
	\begin{abstract}
		Spiking neural networks~(SNNs) have demonstrated excellent capabilities in various intelligent scenarios. Most existing methods for training SNNs are based on the concept of synaptic plasticity; however, learning in the realistic brain also utilizes intrinsic non-synaptic mechanisms of neurons. The spike threshold of biological neurons is a critical intrinsic neuronal feature that exhibits rich dynamics on a millisecond timescale and has been proposed as an underlying mechanism that facilitates neural information processing. In this study, we develop a novel synergistic learning approach that involves simultaneously training synaptic weights and spike thresholds in SNNs. SNNs trained with synapse-threshold synergistic learning~(STL-SNNs) achieve significantly superior performance on various static and neuromorphic datasets than SNNs trained with two degenerated single-learning models. During training, the synergistic learning approach optimizes neural thresholds, providing the network with stable signal transmission via appropriate firing rates. Further analysis indicates that STL-SNNs are robust to noisy data and exhibit low energy consumption for deep network structures. Additionally, the performance of STL-SNN can be further improved by introducing a generalized joint decision framework. Overall, our findings indicate that biologically plausible synergies between synaptic and intrinsic non-synaptic mechanisms may provide a promising approach for developing highly efficient SNN learning methods.
		
	\end{abstract}

	\begin{IEEEkeywords}
		Spiking neural networks, Synergistic learning, Spike threshold, Synaptic plasticity, Joint decision framework.
	\end{IEEEkeywords}
	\IEEEpeerreviewmaketitle

	\section{Introduction}
	\IEEEPARstart{T}{h}e human brain is a complicated organ that is composed of a large number of neurons~\cite{koch1999complexity}. These functionally diverse neurons are fundamental information processing units in the brain and are structurally interconnected in specific configurations by synapses. The human brain has evolved over time and established a wide range of intrinsic mechanisms to ensure its tremendous information processing ability while consuming limited energy~\cite{daoudal2003long, gosak2022networks, klinshov2022rate}. A better understanding of these biologically plausible brain mechanisms is believed to be helpful for designing highly efficient artificial intelligence models~\cite{li2023brain}.

	The spiking neural network~(SNN) is a new generation of neural network model that has recently received considerable attention~\cite{maass1997networks}. In contrast to traditional artificial neural networks~(ANNs), information in SNNs is represented as discrete spikes that are known as action potentials. By continuously integrating inputs, a neuron in an SNN can emit an action potential when its membrane potential crosses a spike threshold, and this action potential will be propagated to postsynaptic neurons. This event-type signal is regarded as the smallest signal unit that can be transmitted between neurons, and it includes information reflecting the inherent dynamics and historically accumulated membrane potential of the neuron. By mimicking brain-inspired computational strategies, SNNs have demonstrated significant potential in both spatio-temporal neural information processing and energy-efficient computation~\cite{pei2019towards, 9665763, 8981937}. Notably, these features enable SNNs to achieve competitive performance in a variety of intelligent tasks and neuromorphic computing applications~\cite{8981937, 8720035}, and can be further applied in broader scenarios such as physical systems~\cite{sigaki2020learning}.

	Recent SNN studies have focused on the development of highly efficient learning methods. However, there is a broad consensus that the direct training of SNNs is considerably more difficult than that of ANNs. To some extent, this result is attributed to the non-differentiable nature of spike activity and the complex neural dynamics in both the spatial and temporal domains. In theory, these limitations make that SNNs cannot be trained directly with the standard backpropagation~(BP) algorithm.  To address these challenges, various learning methods that allow us to perform intelligent tasks with SNNs have been developed. In general, most of these methods focus on the modification of synaptic weights between neurons during the training process and can be roughly classified into four categories: 1) unsupervised learning based on biological synaptic plasticity~\cite{li2012spike, liu2020effective, diehl2015unsupervised, 8354825}; 2) indirect supervised learning via ANN-to-SNN conversions~\cite{wu2021tandem, diehl2015fast}; 3) direct supervised learning including but not limited to spike-based BP methods with surrogate and approximate gradients, reward-based methods, and the FORCE method~\cite{wu2018spatio, wu2019direct, cramer2022surrogate, mozafari2018first, nicola2017supervised, hao2020biologically, 9950361}; and 4) hybrid synergistic approaches that combined different learning strategies~\cite{wu2022brain, zhang2021self, legenstein2008learning}.
	
	Although the training of SNNs has long been dominated by the concept of activity-dependent synaptic modification~\cite{zeng2019short}, synaptic plasticity is not the only learning mechanism in the brain. Recent studies have shown that biological neurons can also adjust their intrinsic excitability to match the dynamic range of synaptic inputs during learning~\cite{zhang2003other, mozzachiodi2010more}.This non-synaptic learning mechanism is referred to as neuronal intrinsic plasticity, and it is assumed to play a functional role in affecting the biophysical properties of neurons at the cellular level~\cite{schrauwen2008improving, stemmler1999voltage, joshi2009rules}. As an important intrinsic neuronal property, the spike threshold of a biological neuron, which can be recorded in different cortical and subcortical areas, is not constant but instead exhibits large variability over a range of a few millivolts both across and within neurons~\cite{azouz1999cellular,  farries2010dynamic, fontaine2014spike, azouz2000dynamic}. The change in the spike threshold has been experimentally observed to occur on a millisecond timescale, and the voltage for spike generation is highly associated with the rate of the preceding membrane depolarization and can be regulated by several intrinsic properties related to neuronal excitability~\cite{fontaine2014spike, azouz2000dynamic}. Further investigations have also revealed that the dynamic spike threshold can significantly enrich the dynamical behaviors of neurons, which may be an underlying mechanism for stabilizing signal transmission and optimizing information processing in the brain~\cite{huang2016adaptive, salaj2021spike}. These findings suggest a possible relation between the dynamic spike threshold and the intrinsic excitability of neurons, implying that the spike thresholds of neurons can be viewed as idealized parameters to be tuned in SNNs.
		
	Recently, several strategies for adjusting SNN threshold-related parameters have been proposed. For unsupervised learning approach, a few information-based rules of intrinsic plasticity have been developed to tune threshold-related parameters in an adaptive manner~\cite{li2012spike, zhang2019information}. For indirect training approaches, various threshold-balancing methods for manipulating the threshold of neurons have been developed~\cite{ding2021optimal, diehl2015fast, sengupta2019going, deng2021optimal}. In these studies, both the threshold-balancing factor and the modified threshold-related activation function are typically introduced to control the spike threshold of neurons in SNNs. With these threshold-balancing methods, the converted SNNs exhibit proper information transmission with appropriate firing rates and can achieve near-lossless accuracy as pre-trained ANNs with equivalent architectures. For direct training approaches, several studies have attempted to capture the rapid dynamics of the spike threshold by an adaptive process of the membrane potential~\cite{salaj2021spike, shaban2021adaptive}. In these studies, the adaptive-related parameters for the spike threshold are treated as hyperparameters. These hyperparameters are fixed for all neurons and remain unchanged during the training process. In principle, these idealized treatments ignore the heterogeneity of biological neurons, thus reducing the highly efficient information processing capabilities that may benefit from neuronal heterogeneity~\cite{perez2021neural}.	
	
	
	In this work, we consider the spike thresholds of neurons to be learnable parameters in SNNs and establish a synergistic learning approach that allows us to simultaneously train the spike thresholds and the synaptic weights in a direct manner. For simplicity, the final SNN trained with the proposed synergistic learning approach is called the ``STL-SNN''. By evaluating our method on both static and neuromorphic benchmark datasets, we show that the STL-SNN model significantly outperforms SNNs trained by single-learning methods~(i.e., the synaptic learning~(SL) method and the threshold learning~(TL) method). In particular, our detailed analysis reveals that suitable synergies between spike thresholds and synaptic weights endow SNNs with strong noise robustness, stable signal transmission and reasonable energy consumption. Furthermore, we introduce a simple yet effective joint decision framework for SNNs to prevent potential decision difficulties caused by the rate-based decoding scheme. Our results thus emphasize the importance of brain-inspired synergistic information processing mechanisms in developing highly efficient SNN models.
	
	The main contributions and highlights of this study can be summarized as follows:
	\begin{itemize}
		\item[$\bullet$] We propose a direct synapse-threshold synergistic learning approach for SNNs at the single-neuron level. With this approach,  the synaptic weights and thresholds of neurons can be trained simultaneously in SNNs based on the spike-based BP method.
		\item[$\bullet$] Our results suggest that appropriate synergies between synaptic weights and spike thresholds are critical for training high-performance SNNs, thus endowing SNNs with superior robustness, stability, and energy consumption.
		\item[$\bullet$] The experimental results show that our developed STL-SNN model can achieve better or competitive performance compared to other state-of-the-art models on mainstream various benchmark datasets with different tasks.
		\item[$\bullet$] An efficient joint decision framework is proposed to address the decision difficulties for SNNs with rate-based decoding schemes.
	\end{itemize}

	
	The remainder of this paper is organized as follows. In Section II, a brief introduction on some related works on direct supervised learning and hybrid synergistic learning approaches for SNNs is provided. In Section III, we describe the proposed STL-SNN model and its learning algorithm in detail. In Section IV, we systematically present the datasets, network structures, experimental results and joint decision framework. Finally, detailed discussions and conclusions are presented in Section V.

	\section{Related Work}
	Our work aims to design a novel approach for simultaneously training the synaptic weights and spike thresholds in SNNs at the single-neuron level. To this end, we develop a synapse-threshold synergistic learning method for directly training SNNs in a spike-based BP framework. In the following section, we briefly review several recent works on spike-based BP methods and hybrid synergistic learning strategies for SNNs.

	The BP algorithm is a powerful supervised learning approach for traditional ANN models~\cite{lecun2015deep}. However, it has been widely acknowledged that directly training SNNs with the standard BP method is quite difficult, because the gradient information with respect to the loss function cannot be calculated easily due to the non-differentiable nature of spiking events. To address this challenge, several spike-based BP methods that utilize surrogate and approximate gradients have been proposed~\cite{wu2018spatio, cramer2022surrogate}. These direct training methods for SNNs include, but are not limited to, the hybrid macro-micro BP approach~\cite{jin2018hybrid}, the spatio-temporal BP~(STBP) approach~\cite{wu2018spatio, wu2019direct}, a BP approach using SLAYER~\cite{shrestha2018slayer}, and the spike-train level recurrent SNN BP~(ST-RSBP) approach~\cite{zhang2019spike}. Among them, the STBP approach is believed to be a promising algorithm,  which are capable of achieving high accuracy on typical static and neuromorphic datasets with a relatively short time window due to its outstanding spatio-temporal neural information processing capability. Recently, several normalization techniques have been also integrated into the STBP method to further improve its performance in training SNNs. For instance, a previous study incorporated a neuron normalization (NeuNorm) strategy to balance neural selectivity, confirming that the STBP method with NeuNorm can be applied to train deeper and larger SNNs with improved performance~\cite{wu2019direct}. In the next section, we develop the synergistic learning rule with the STBP method by considering its advantages in both efficiency and flexibility.
	
	Remarkably, another popular vein for enhancing the training capabilities of SNNs is to develop efficient hybrid synergistic learning strategies. Several hybrid synergistic learning models have recently been proposed, and these methods have demonstrated superior performance to single-learning models. In general, hybrid synergistic learning can be developed by combining various learning algorithms, and a commonly used approach is to incorporate biologically-inspired local learning with supervised error- or reward-based global learning~\cite{wu2022brain, zhang2021self}. Despite significant progress, the majority of existing hybrid synergistic learning strategies focus on the modification of synaptic weights during the learning process. In addition to synaptic plasticity, experimental evidence has indicated that learning in the brain also benefits greatly from several intrinsic non-synaptic mechanisms~\cite{zhang2003other, mozzachiodi2010more}. However, only limited works have investigated hybrid synergistic learning models for SNNs by incorporating synaptic and intrinsic plasticity~\cite{salaj2021spike, huang2016adaptive}. As mentioned above, previous experimental studies have established a relation between the dynamic spike threshold and the intrinsic excitability of neurons~\cite{azouz1999cellular,  farries2010dynamic, fontaine2014spike, azouz2000dynamic}. Inspired by these findings, in this work, we introduce the concept of a learnable spike threshold and present a highly effective synergistic learning approach that simultaneously trains synaptic weights and spike thresholds in SNNs.
	
	\begin{figure*}[htbp]
		\centering
		\includegraphics[width=17.6cm]{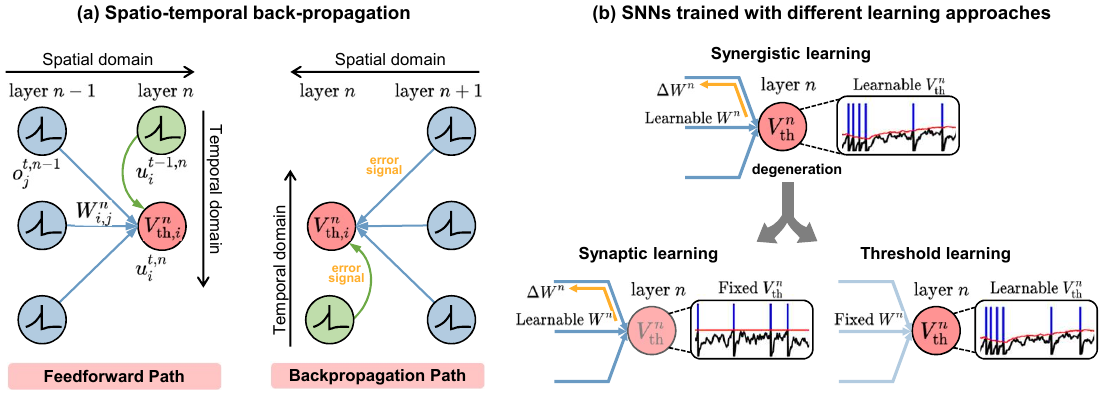} 
		\caption{The spatio-temporal back-propagation and learning methods in SNNs. (a) Schematic diagram of spatio-temporal back-propagation. In the feedforward path, each spiking neuron inherits its previous information in the temporal domain and receives inputs from the preceding layer in the spatial domain. In the backpropagation path, the error signal is propagated in both the spatial and temporal domains. (b) An illustration of SNNs trained with three different learning approaches, including synergistic learning between synaptic weights and spike thresholds and two degenerate single-learning versions based on synaptic learning and threshold learning.}
		\label{Fig1}
	\end{figure*}
	
	\section{Methods}
	In this section, we first describe the leaky-integrate-and-fire~(LIF) neuron model, as well as its iterative form used in this work. Then, we mathematically derive the proposed synapse-threshold synergistic learning algorithm for directly training SNNs within the framework of the STBP method using approximate gradients~\cite{wu2018spatio}.

	\subsection{LIF Neuron Model}
    To date, a variety of spiking neuron models that simulate the dynamics of biological neurons in the brain have been proposed~\cite{CHENG2023217}. The LIF model is a typical neuron model used in SNNs that processes rich spatio-temporal neuronal information in a simple yet efficient way~\cite{gerstner2014neuronal}. Mathematically, the dynamics of the LIF neuron can be described as follows:
    
	\begin{equation}
	\label{Equation1}
		\tau \frac{du^t}{dt}=  V_{\text{rest}} -u^t + I^t
	\end{equation}
	with the spiking generation mechanism
	\begin{equation}
	\label{Equation2} 
		o^t =
			\begin{cases}
				1  & \mbox{if }u^t\ge V_{\text{th}},\\
				0  & \text{otherwise},
			\end{cases}
	\end{equation}
	where $u^t$ and $I^t$ represent the membrane potential and input of the LIF neuron at time $t$, respectively, and $\tau$ is the time constant of the membrane potential. When  $u^t$ exceeds the spike threshold $V_{\text{th}}$, a spike is generated and the membrane potential is reset to the resting potential $V_{\text{rest}}$. For simplicity, we set $V_{\text{rest}}=0$~mV and provide the discrete form of the differential equation shown in Eq.~(\ref{Equation1}) using approximate iterations:
	\begin{equation}
	\label{Equation3}
		u^t = \alpha^t \cdot  u^{t-1} + \frac{dt}{\tau}I^t.
	\end{equation}
	In Eq.~(\ref{Equation3}), the variable $\alpha^t$ is the decay factor of the membrane potential in the temporal domain. Theoretically, this variable depends on the binary spike train and can be described as follows: 
	\begin{equation}
	\label{Equation4} 
	\alpha^t=
		\begin{cases}
			1-\frac{dt}{\tau} & \mbox{if } o^{t-1}=0, \\
			0                 & \mbox{if } o^{t-1}=1.
		\end{cases}
	\end{equation}

	\subsection{SNN with Synapse-Threshold Synergistic Learning}
	We begin by introducing the synapse-threhsold synergistic learning approach for SNNs. To doing so, we employ the STBP method to derive the proposed synergistic learning algorithm~\cite{wu2018spatio}. Fig.~\ref{Fig1}(a) shows the feedforward and backpropagation paths in the spatial and temporal domains, respectively.
	
	In the feedforward path, each neuron in the SNN inherits its historical information with a decay factor in the temporal domain and receives neuronal inputs from the preceding layer in the spatial domain [Fig.~\ref{Fig1}(a), left]. Without loss of generality, the dynamical evaluation of the membrane potential of a neuron in an SNN can be written in the following iterative form:
	\begin{equation}
	\label{Equation5}
	u_{i}^{t, n}  = \alpha_{i}^{t, n} \cdot u_{i}^{t-1, n}+ \frac{dt}{\tau}I_{i}^{t, n} ,  
	\end{equation}
	where $u_{i}^{t, n}$ is the membrane potential of the $i$-th neuron in the $n$-th layer at time $t$, and $\alpha_{i}^{t, n}$ and $I_{i}^{t, n}$ denote the corresponding decay factor and total input of the neuron, respectively. The binary output of the neuron depends on both the membrane potential and the spike threshold:
	\begin{equation} 
	\label{Equation6}
	o_{i}^{t, n}  = \sigma(u_{i}^{t, n}, V_{\text{th}, i}^{n}),
	\end{equation}
	where  $V_{\text{th}, i}^{n}$ is the spike threshold of the $i$-th neuron in the $n$-th layer, and	$\sigma$ is an approximate gradient function that addresses the non-differentiability of the spike activity. By incorporating the scaling effect $\frac{dt}{\tau}$ into the synaptic weights, we can formulate the scaled input as follows:
	\begin{equation}
	\label{Equation7}
	x_{i}^{t, n} = \frac{dt}{\tau} I_{i}^{t, n} = \sum_j w_{ij}^{n} o_{j}^{t, n-1}.   
	\end{equation}
	Here, the outer sum runs over all the synapses onto the $i$-th neuron in the $n$-th layer, $w_{ij}^n$ represents the scaled synaptic weights from the $j$-th neuron in the preceding layer and $W_{i}^n=[w_{i1}^n, w_{i2}^n, ..., w_{il(n-1)}^n]$ is the vector form of the synaptic weights.
	
	To learn both the synaptic weights and the spike thresholds in an SNN, we employ the rate-based decoding scheme~\cite{wu2019direct} and define the mean square error (MSE) as the loss function in this work. The MSE loss function can be described as:
	\begin{equation}
	\label{Equation8}
	L_{\text{MSE}}=\frac{1}{2S} \sum_{s=1}^{S} \sum_{c=1}^{C}\left [ y_{c,s}-\frac{1}{TP}\sum_{i\in c}\sum_{t=1}^{T}o_{i,s}^{t, N} \right]^2.
	\end{equation}
	We set the last layer as a voting layer and each class is represented by one neural population~(the total number of classes is $C$ and the size of each neural population is $P$). $y_{c,s}$ is the label of the input samples and S is the number of training samples. By measuring the MSE between the average voting results and the label, the requirement for a long time window can be alleviated~\cite{wu2019direct, fang2021incorporating}.

	In this study, we use the STBP method to calculate the gradients and update the learnable SNN parameters to minimize the loss function~\cite{wu2018spatio}. In the STBP method, the derivatives of the loss function with respect to $ o_{i}^{n} (t)$ and $u_{i}^{n}(t)$ depend on the error backpropagation in both the spatial and temporal domains~[Fig.\ref{Fig1}~(a), right]. According to the chain rule, the derivatives $\frac{\partial L_{\text{MSE}}}{\partial o_{i}^{n} (t)}$ and $\frac{\partial L_{\text{MSE}}}{\partial u_{i}^{n}(t)}$ can be mathematically described as follows~\cite{wu2018spatio}:
	\begin{equation}
	\label{Equation9}
	\frac{\partial L_{\text{MSE}}}{\partial o_{i}^{t,n}} = \frac{\partial L_{\text{MSE}}}{\partial o_{i}^{t+1, n}} \frac{\partial o_{i}^{t+1, n}}{\partial o_{i}^{t, n}}+\sum_{j=1}^{l(n+1)}\frac{\partial L_{\text{MSE}}}{\partial o_{j}^{t, n+1}}\frac{\partial o_{j}^{t, n+1}}{\partial o_{i}^{t, n}},
	\end{equation}
	\begin{equation}
	\label{Equation10}
	\frac{\partial L_{\text{MSE}}}{\partial u_{i}^{t, n}} =\frac{\partial L_{\text{MSE}}}{\partial o_{i}^{t, n}} \frac{\partial o_{i}^{t, n}}{\partial u_{i}^{t, n}}
	+\frac{\partial L_{\text{MSE}}}{\partial o_{i}^{t+1, n}} \frac{\partial o_{i}^{t+1, n}}{\partial u_{i}^{t, n}}  .
	\end{equation}
	
	Based on Eqs.~(\ref{Equation9}) and~(\ref{Equation10}), we finally obtain the derivatives with respect to the synaptic weights and the spike thresholds of neurons in the $n$-layer as follows:
	\begin{equation}
	\label{Equation11}
	\frac{\partial L_\text{MSE}}{\partial W^{n}}=\sum_{t=1}^{T}\frac{\partial L_{\text{MSE}}}{\partial u^{t, n}} \frac{\partial u^{t, n}}{\partial x^{t, n}} \frac{\partial x^{t, n}}{\partial W^{n}},
	\end{equation}
	\begin{equation}
    \label{Equation12}
    \frac{\partial L_{\text{MSE}}}{\partial V_{\text{th}}^{n}} =\sum_{t=1}^{T}\frac{\partial L_{\text{MSE}}}{\partial o^{t, n}} \frac{\partial o^{t, n}}{\partial V_{\text{th}}^{n}},
    \end{equation}
  	where  $u^{t, n}$, $x^{t, n}$ and $o^{t, n}$ are the vector forms of the membrane potential, inputs and spike events of neurons in the $n$-th layer at time $t$, respectively, $W^n=[W_{1}^n, W_{2}^n, ..., W_{l(n)}^n]^\text{T}$ represents the matrix of synaptic weights from the $(n-1)$-th layer to the $n$-th layer, and $V_{\text{th}}^{n}=[V_{\text{th, 1}}^{n}, V_{\text{th, 2}}^{n}, ..., V_{\text{th},~l(n)}^{n}]^\text{T}$ denotes the spike thresholds of neurons in the $n$-th layer .

    To address the challenge of the non-differential nature of the spiking events in SNNs, we choose the approximate gradient function with the following form~\cite{wu2018spatio, fang2021incorporating}: 
	\begin{equation}\label{Equation13}
	\sigma(u_{i}^{t, n}, V_{\text{th}, i}^{n}) = \frac{\text{arctan} \left [ \pi (u_{i}^{t, n} - V_{\text{th}, i}^{n}) \right ] }{\pi}+\frac{1}{2}.
	\end{equation}
	
	In additional experiments, we demonstrate that similar results can be observed with different approximate gradient functions~\cite{wu2018spatio} .
	
	By using Eqs.~(\ref{Equation8})--(\ref{Equation13}, we can simultaneously train both the synaptic weights and thresholds of neurons in SNNs at the single-neuron level. This means that each neuron has an individual learnable threshold in the STL-SNN model. Indeed, our design philosophy is clear and bio-constrained, as the spike threshold is an important intrinsic property of biological neurons, and different neurons have their own thresholds in the brain. Note that the SNN trained by our synergistic learning approach~(STL-SNN) can be degenerated into a single-learning model when either the spike thresholds or synaptic weights are set as constants in the network [Fig.~\ref{Fig1}(b)]. These two single-learning models can be mathematically represented as $\frac{\partial L_{\text{MSE}}}{\partial V_{\text{th}, i}^{n}}=0$ and $\frac{\partial L_\text{MSE}}{\partial W_{ij}^{n}}=0$, corresponding to an SNN trained with synaptic learning~(SL-SNN) and threshold learning~(TL-SNN), respectively. The source codes of our SNN models will be provided on GitHub (https://github.com/GuoLab-UESTC) after the acceptance of this manuscript.

	    \begin{figure*}[htbp]
		\centering
		\includegraphics[width=15.6cm]{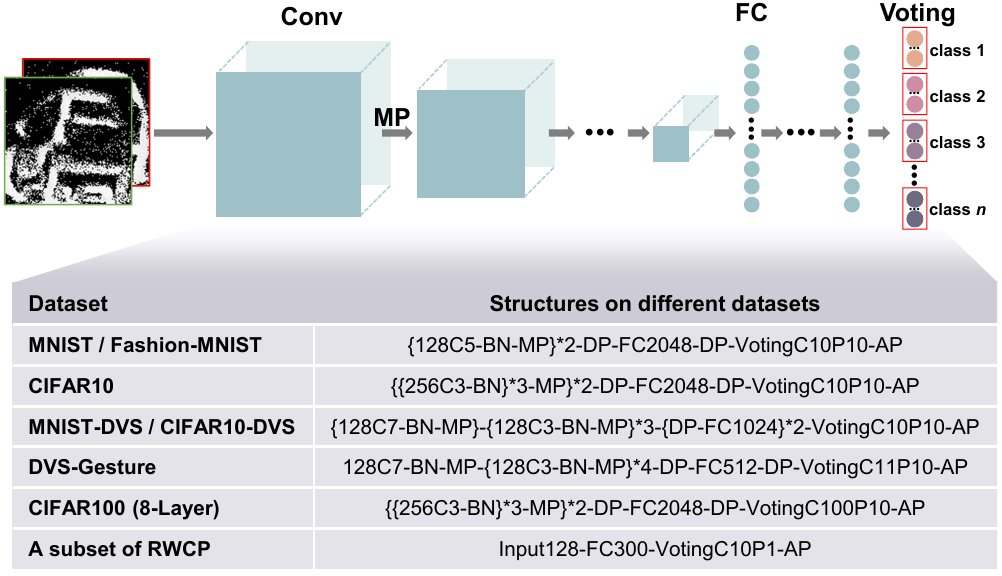} 
		\caption{Network structure used in this study. The upper panel schematically shows the structure of SNNs, which includes convolutional layers (Conv), fully connected layers (FC) and a voting layer (Voting). In addition, the batch normalization (BN) and max pooling (MP) techniques are applied to the Conv layers, and dropout (DP) with $p=0.5$ is used for the FC layers. The bottom panel illustrates the detailed structures for the different datasets. For example, a 4-Layer SNN with $\left  [\text{\{128C3-BN-MP\}*2-DP-FC2048-DP-VotingC10P10-AP}\right]$ is trained on MNIST, where 128C3 represents the Conv layers with 128 output channels and a kernel size of $3\times 3$, FC2048 indicates the FC layer with 2048 output features, VotingC10P10 indicates 10 classes in the voting layer with 10 neurons for each class, and the symbol $\{\cdot\}*$ indicates the repeated structure of the network.}
		\label{Fig2} 
	\end{figure*} 

	\section{Experiments and Results}
		
	\subsection{Datasets and Networks}
	To evaluate the performance of SNNs trained with different learning methods, we conduct most of the experiments on both static datasets~(MNIST, Fashion-MNIST and CIFAR10) and neuromorphic datasets~(MNIST-DVS, CIFAR10-DVS and DVS-Gesture). For the MNIST dataset, we use the Bernoulli generator to convert the pixels in the images to spike trains with a time window of $4$~ms to use as the input to the SNN. For the Fashion-MNIST and CIFAR10 datasets, which include more complex scenarios, the input layers are considered the encoding layer that directly receives real pixels in the images, and the time windows are all set to $8$~ms. Since the default training and test sets of these static datasets are provided, we employ these sets as the defaults for evaluating the performance of the SNNs. For CIFAR10-DVS and DVS-Gesture neuromorphic datasets with complicated spatio-temporal neural dynamics, we divide the events of each data into 20 slices with nearly the same number of events in each slice and integrate events to frames~\cite{fang2021incorporating}. For the simple MNIST-DVS dataset, we only use the first 100~ms of data and accumulate raw data into frames with a fixed slice length of 5~ms. There are three different recording scales in the MNIST-DVS dataset, and the scale of 4 is used in this work. We feed these preprocessed neuromorphic data (corresponding to a given time window of $T=20$~ms) into the SNNs for training and testing. For the different neuromorphic datasets, the sizes of the training and test sets are chosen as follows: 9000:1000 (MNIST-DVS), 9000:1000 (CIFAR10-DVS) and 1056:288 (DVS-Gesture).

	Moreover, to further examine the effectiveness of our proposed synapse-threshold synergistic learning method, we also conduct several experiments on more challenging and complicated datasets, including the CIFAR100 and RWCP speech datasets. The CIFAR100 dataset consists of 100 categories, with each category containing 600 static images. The default numbers of training and test samples per category are 500 and 100, respectively. In this study, we employ the same data preprocessing strategy as for the CIFAR10 dataset to obtain the inputs to SNNs. The RWCP dataset is composed of isolated sound events~\cite{nakamura2000acoustical}, and we use a subset for unsegmented sensory event detection created in a previous work in experiments~\cite{gu2019stca}. To facilitate data encoding, we divide the events of each sample into 50 slices and then accumulate them into frames~\cite{gu2019stca}.

	As shown in Fig.~\ref{Fig2} (upper panel), the structure of the SNNs established in our study is composed of the convolutional layers (Conv), fully connected layers (FC) and a population voting layer (Voting). In addition, the batch normalization (BN), dropout (DP) and max pooling (MP) techniques are introduced into the SNNs to prevent network overfitting and downsample the feature maps. A rate-based decoding scheme is used to distinguish different classification categories~\cite{wu2019direct}. To accomplish this, we configure a voting layer with several neural populations and each output class is denoted by one neural population. During the inference phase, the total spike events for the neurons in each output class are calculated, and the output class with the highest spiking activity is considered as the SNN classification result~\cite{wu2019direct}. We design networks with different configurations for the different datasets, and the detailed information is presented in Fig.~\ref{Fig2} (bottom panel). Unless otherwise stated, we use the default values of the hyperparameters listed in Tab.~\ref{Tab1} for the different datasets. In all experiments, the learning rate is decayed in an exponential manner with two hyperparameters (initial learning rate $\eta$ and decay factor $\gamma$) during the training process.
	
	\begin{figure*}[htbp]
		\centering
		\includegraphics[width=18.1cm]{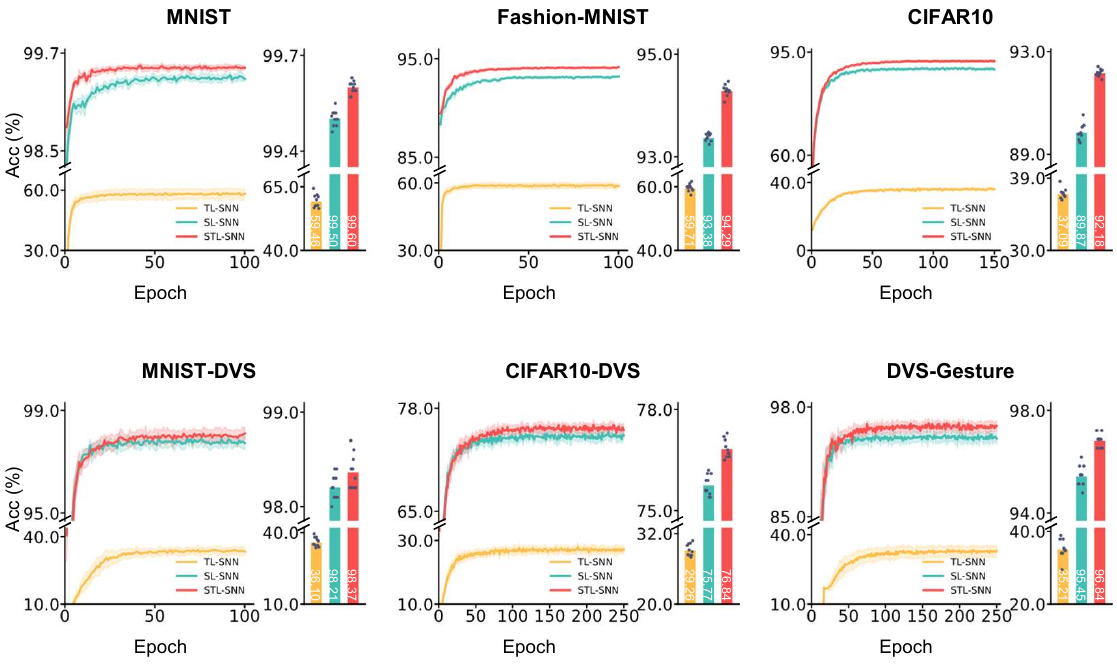} 
		\caption{Performance of SNNs trained with different learning approaches on both static and neuromorphic datasets. All experiments are repeated for 10 trials. On each dataset, the average training curves with standard deviations for the various learning approaches are presented (left). For a detailed comparison, both the top-1 accuracy of each trial (dot) and the average top-1 accuracy (histogram) are also plotted (right). On different datasets, the STL-SNN model exhibits better performance than the TL-SNN and SL-SNN models.}
		\label{Fig3} 
	\end{figure*}

\begin{table*}[htbp]
	\centering
	\caption{Detailed spiking neural networks hyperparameters for different datasets.}
	\label{Tab1}
	\renewcommand\arraystretch{1.2}
	\setlength{\tabcolsep}{0.5mm}
	\begin{tabular}{l|m{1.4cm}<{\centering}m{1.4cm}<{\centering}m{1.4cm}<{\centering}m{1.4cm}<{\centering}m{1.4cm}<{\centering}m{1.4cm}<{\centering}m{1.4cm}<{\centering}m{1.4cm}<{\centering}m{1.4cm}<{\centering}}
		\hline
		\textbf{Hyper-parameters}               & \textbf{MNIST} & \textbf{Fashion-MNIST} & \textbf{CIFAR10} & \textbf{MNIST-DVS} & \textbf{CIFAR10-DVS} & \textbf{DVS-Gesture} & \textbf{CIFAR100 ~(8-Layer)} & \textbf{CIFAR100 ~(ResNet18)} & \textbf{RWCP}\\ 
		\hline
		Initial threshold                       & 2.0 mV         & 2.0 mV       & 2.0 mV           & 2.0 mV         & 2.0 mV           & 2.0 mV     & 2.0 mV   & 1.0 mV    & 2.0 mV \\
		Time constant $\tau$                    & 2.0 ms         & 2.0 ms       & 2.0 ms           & 2.0 ms         & 2.0 ms           & 2.0 ms     & 2.0 ms   & 2.0 ms    & 2.0 ms \\
		Time window $T$                         & 4 ms           & 8 ms         & 8 ms             & 20 ms          & 20 ms            & 20 ms      & 8 ms     & 4 ms      & 50 ms  \\
		Time step $dt$                          & 1 ms           & 1 ms         & 1 ms             & 1 ms           & 1 ms             & 1 ms       & 1 ms     & 1 ms      & 1 ms   \\
		Batch size                              & 50             & 50           & 40               & 40             & 40               & 16         & 40       & 200       & 10     \\
		Number of training epochs               & 100            & 100          & 150              & 100            & 250              & 250        & 400      & 400       & 100 \\
		Initial learning rate $\eta$            & 0.001          & 0.001        & 0.001            & 0.001          & 0.001            & 0.001      & 0.001    & 0.01      & 0.001 \\
		Decay factor of learning rate $\gamma$  & 0.93           & 0.93         & 0.95             & 0.93           & 0.97             & 0.97       & 0.98     & 0.98      & 0.93\\
		Dropout rate $p$                        & 0.50           & 0.50         & 0.50             & 0.50           & 0.50             & 0.50       & 0.50     & --        & -- \\
		\hline
	\end{tabular}
\end{table*}

	 \begin{table*}[htbp]
		\centering
		\caption{Detailed comparison of SNNs trained with different learning approaches in the present study. Each experiment is repeated for 10 trials with different random seeds. The average top-1 accuracy in the form of the mean $\pm$ standard deviation~(upper) and the best top-1 accuracy of 10 trials (bottom) are presented.}
		\label{Tab2}
		\renewcommand\arraystretch{1.2}
		\begin{tabular}{l|p{2cm}<{\centering}p{2cm}<{\centering}p{2cm}<{\centering}p{2cm}<{\centering}p{2cm}<{\centering}p{2cm}<{\centering}}
			\hline
			\textbf{Dataset}                        & \textbf{TL-SNN}   & \textbf{SL-SNN}   & \textbf{STL-SNN} & \textbf{Hete-SNN}   & \textbf{SL-SNN (JDF)}   & \textbf{STL-SNN (JDF)}   \\ 
			\hline
			\multirow{2}{*}{\textbf{MNIST}}         & 59.46$\pm$2.51\%  & 99.50$\pm$0.02\%  & 99.60$\pm$0.02\% & 99.54$\pm$0.03\%    & 99.55$\pm$0.02\%        & 99.62$\pm$0.01\%                    \\
													& 64.39\%           & 99.55\%           & 99.63\%          & 99.58\%             & 99.58\%                 & 99.64\%     \\   	 \hline
			\multirow{2}{*}{\textbf{Fashion-MNIST}} & 59.71$\pm$1.15\%  & 93.38$\pm$0.07\%  & 94.29$\pm$0.10\% & 93.49$\pm$0.11\%    & 93.67$\pm$0.07\%        & 94.53$\pm$0.09\%                    \\
													& 61.68\%           & 93.48\%           & 94.47\%          & 93.76\%             & 93.77\%                 & 94.68\%     \\      \hline 
			\multirow{2}{*}{\textbf{CIFAR10}}       & 37.09$\pm$0.61\%  & 89.87$\pm$0.32\%  & 92.18$\pm$0.14\% & 87.30$\pm$0.31\%    & 91.13$\pm$0.25\%        & 93.25$\pm$0.09\%                    \\
													& 38.59\%           & 90.54\%           & 92.42\%          & 87.68\%             & 91.45\%                 & 93.38\%     \\      \hline
			\multirow{2}{*}{\textbf{MNIST-DVS}}     & 36.10$\pm$1.89\%  & 98.21$\pm$0.13\%  & 98.37$\pm$0.17\% & 98.36$\pm$0.24\%    & 98.34$\pm$0.07\%        & 98.55$\pm$0.15\%                    \\
													& 39.50\%           & 98.40\%           & 98.70\%          & 98.70\%             & 98.50\%                 & 98.80\%      \\     \hline
			\multirow{2}{*}{\textbf{CIFAR10-DVS}}   & 29.26$\pm$0.98\%  & 75.77$\pm$0.29\%  & 76.84$\pm$0.26\% & 73.38$\pm$1.61\%    & 76.85$\pm$0.39\%        & 77.82$\pm$0.32\%                    \\
													& 30.80\%           & 76.20\%           & 77.30\%          & 76.10\%             & 77.4\%                  & 78.5\%       \\     \hline
			\multirow{2}{*}{\textbf{DVS-Gesture}}   & 35.21$\pm$2.51\%  & 95.45$\pm$0.39\%  & 96.84$\pm$0.24\% & 95.57$\pm$0.38\%    & 95.76$\pm$0.40\%        & 97.01$\pm$0.23\%                    \\
													& 38.89\%           & 96.18\%           & 97.22\%          & 96.18\%             & 96.52\%                 & 97.22\%      \\     \hline
		\end{tabular}
	\end{table*}

	\begin{table*}[b]
	\centering
	\caption{\label{Tab3} Comparison with existing state-of-the-art results on both static and neuromorphic datasets. }
	\renewcommand\arraystretch{1.2}
	\begin{tabular}{lccccccc}
		\hline
		Model              				& Method         & MNIST         & Fashion-MNIST   & CIFAR10      & MNIST-DVS   & CIFAR10-DVS   & DVS-Gesture   \\ \hline
		STBP~\cite{wu2018spatio}      	& Spike-based BP & 99.42\%       & --              & 50.70\%      & --          & --            & --            \\
		NeuNorm~\cite{wu2019direct}    	& Spike-based BP & --            & --              & 90.53\%      & --          & 60.50\%       & --            \\
		SPA~\cite{liu2020effective}      & Spike-based BP & --           & --              & --           & --          & 32.20\%       & --            \\
		PLIF~\cite{fang2021incorporating}& Spike-based BP & \textbf{99.72\%}      & 94.38\%         & \textbf{93.50\%}      & --          & 74.80\%       & \textbf{97.57\%}       \\
		TSSL-BP~\cite{zhang2020temporal} & Spike-based BP & 99.53\%      & 92.83\%         & 91.41\%      & --          & --            & --            \\
		ST-RSBP~\cite{zhang2019spike}    & Spike-based BP & 99.62\%      & 90.13\%         & --           & --          & --            & --            \\
		HP~\cite{wu2022brain}          	& Spike-based BP  & 99.50$\pm$0.04\%  & 93.29$\pm$0.07\%    & 91.08$\pm$0.09\% & --          & 67.81$\pm$0.34\%  & \textbf{97.01$\pm$0.21\%}  \\
		SLAYER~\cite{shrestha2018slayer} & Spike-based BP & 99.36$\pm$0.05\% & --              & --           & --          & --            & 93.64$\pm$0.49\%  \\
		SBP~\cite{zhang2021self}         & Spike-based BP & 97.89$\pm$0.08\% & --              & --           & --          & --            & 70.88$\pm$0.82\%  \\ \hline
		STBP-tdBN~\cite{zheng2020going}  & ANN2SNN        & --           & --              & 93.16\%      & --          & 67.80\%       & 96.87\%       \\
		TandomSNN~\cite{wu2021tandem}    & ANN2SNN        & --           & --              & 82.78\%      & --          & 63.73\%       & --            \\
		Spike-Norm~\cite{sengupta2019going}  & ANN2SNN    & --           & --              & 91.55\%      & --          & --            & --            \\ \hline
		HATS~\cite{sironi2018hats}       & HATS           & --           & --              & --           & 98.4\%      & 52.40\%       & --            \\
		RG-CNN~\cite{bi2020graph}        & DNN            & --           & --              & --           & 98.6\%      & 54.00\%       & --            \\
		DART~\cite{ramesh2019dart}       & DNN            & --           & --              & --           & --          & 65.78\%       & --            \\ 
		AER~\cite{liu2020unsupervised}	 & STDP           & --           & --              & --           & 89.96\%     & --            & 95.75\%       \\ \hline
		STL-SNN (Best top-1)           				& Spike-based BP & 99.63\%      & \textbf{94.47\%}         & 92.42\%      & \textbf{98.70\%}     & \textbf{77.30\%}       & 97.22\%        \\
		STL-SNN  (Average top-1)          				& Spike-based BP & \textbf{99.60$\pm$0.02\%} & \textbf{94.29$\pm$0.10\%}    & \textbf{92.18$\pm$0.14\%} & \textbf{98.37$\pm$0.17\%} & \textbf{76.84$\pm$0.26\%}  & 96.84$\pm$0.24\%   \\ \hline
	\end{tabular}
	\end{table*}

	\begin{figure*}[t]
		\centering
		\includegraphics[width=18.2cm]{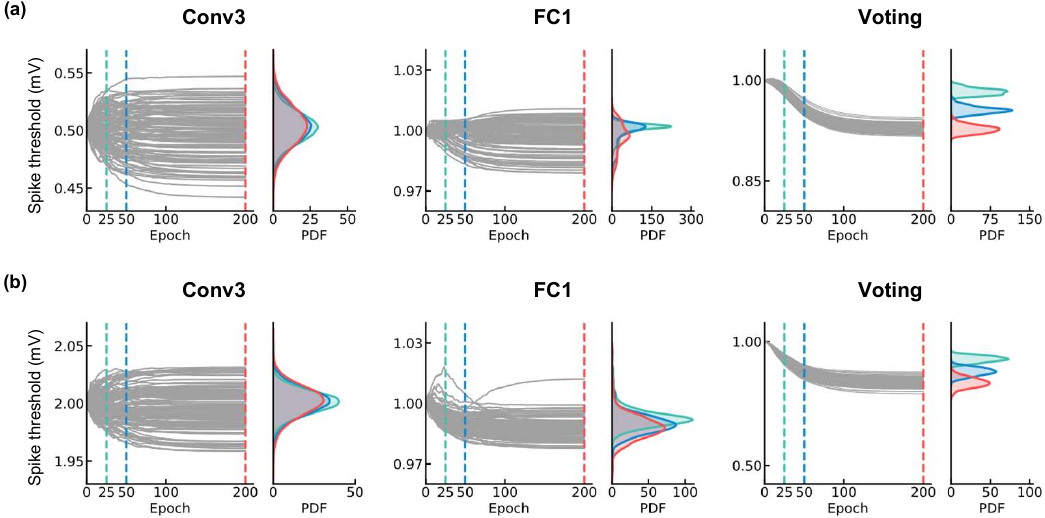} 
		\caption{Evolution of spike thresholds in STL-SNNs trained with two different initial conditions. Here, we set the thresholds for the neurons in the convolutional layers to $0.5$~mV (a) and  $2.0$~mV (b), and the initial spike thresholds for other neurons are all fixed to $1.0$~mV. For simplicity, we only display the threshold evolution in the third convolutional layer (Conv3), the first fully connected layer (FC1) and the voting layer (Voting). In each layer, both the training curves of the spike thresholds for 100 randomly selected neurons and the probability density function (PDF) of all spike thresholds at three specific moments~(epoch = 25, 50, 200) are presented, respectively. After 100 training epochs, the thresholds of neurons in the STL-SNN model converge to an approximately steady state.}
		\label{Fig4} 
	\end{figure*}

	\subsection{Performance Evaluation of SNNs with Synapse-Threshold Synergistic Learning}
	\subsubsection{Ablation Study and Comparison with State-of-the-Art}
	To comprehensively evaluate the performance of the porposed synapse-threshold synergistic learning approach, we conduct an ablation study on a variety of static and neuromorphic benchmark datasets and compare the proposed model (STL-SNN) with two single-learning models, i.e., SL-SNNs and TL-SNNs. In Fig.~\ref{Fig3}, we show the average training curves (left) and both the average top-1 accuracy and the top-1 accuracies of all trials (right) for different datasets. Although both single-learning models exhibit certain capabilities for training SNNs, the SL-SNN model has significantly higher accuracies than the TL-SNN model on the different datasets. From the theoretical viewpoint, this result is not surprising because the number of synapses is considerably greater than the number of neural spike thresholds, providing the SL-SNN model with a superior performance. Remarkably, compared with the two single-learning models, the STL-SNN model achieves higher accuracies on all datasets used in this work. Specifically, with the STL-SNN model, the average top-1 accuracy increases by greater than or close to 1\% on several complicated datasets (0.91\% on Fashion-MINST, 2.31\% on CIFAR10, 1.07\% on CIFAR10-DVS and 1.39\% on DVS-Gesture, Tab.\ref{Tab2}) and by slightly more than 0.1\% on simpler datasets (0.1\% on MNIST and 0.16\% on MNIST-DVS). Moreover, the distribution of the top-1 accuracies of different trials [see the dots on the histograms in Fig.~\ref{Fig3}] demonstrates the stability of the STL-SNN model, with most datasets having a relatively smaller standard deviations of top-1 accuracies than those obtained with the two single-learning models [Tab.~\ref{Tab2}]. Overall, our above results indicate that the synergistic learning approach outperforms single-learning approaches on both static and neuromorphic datasets.

    Furthermore, we compare the accuracies of SNNs trained by our proposed synapse-threshold synergistic learning approach with other state-of-the-art methods, including SNNs trained by spike-based BP methods, SNNs converted from ANNs~(ANN-to-SNN), deep neural networks (DNNs), and statistics-based methods. As shown in Tab.~\ref{Tab3}, the STL-SNN model exhibits the most outstanding performance, with both the best top-1 accuracy and average top-1 accuracy on three datasets, including the Fashion-MNIST (94.47\%, 94.29$\pm$0.10\%), MNIST-DVS (98.70\%, 98.37$\pm$0.17\%) and CIFAR10-DVS (77.30\%, 76.84$\pm$0.26\%) datasets. On the other three datasets, our results are slightly lower than the state-of-the-art accuracies mainly reported in a previous study~\cite{fang2021incorporating}. However, compared with the results of~\cite{fang2021incorporating}, we demonstrate that the STL-SNN model can achieve a competitive performance with fewer training epochs. To some extent, this result may further indicate that appropriate synergies between synaptic weights and spike thresholds can accelerate the convergence speed, thus endowing spiking neural networks with faster learning abilities. 

	In the following of this subsection, we analyze several superior features of the STL-SNN model in detail. Considering the low performance of the TL-SNN model, we mainly compare the results of the STL-SNNs and SL-SNNs on different static and neuromorphic benchmark datasets.
	
	\subsubsection{Evolution of Spike Thresholds in STL-SNNs}
  	To examine how the spike thresholds dynamically change during the learning process, we train two STL-SNNs with different initial thresholds on the CIFAR10-DVS dataset. We initialize the thresholds for neurons in the convolutional layers with different values ($0.5$~mV for the first STL-SNN model and $2.0$~mV for the second STL-SNN model). For simplicity, the initial spike thresholds of the other neurons in these two STL-SNNs are fixed to $1.0$~mV. In addition, the network structure and the other hyperparameters are the same as our previous experiments.
    
    In Fig.~\ref{Fig4}, we plot the evolution curves of the spike thresholds for 100 randomly sampled neurons in the third convolutional layer, the first fully connected layer and the voting layer, respectively. For the two different initial conditions, we consistently observe that the spike thresholds of neurons in different layers quickly adjust during the early stages of training and almost converge to their steady values after 100 epochs of training. This dynamic threshold evolution roughly matches the training curve of the STL-SNN model trained on the CIFAR10-DVS dataset, as depicted in Fig.~\ref{Fig3}. After the task is learned, we observe that the spike thresholds of neurons in different layers have broadly heterogeneous distributions [see the red distributions in Fig.~\ref{Fig4}]. These results clearly indicate that suitable thresholds can be automatically obtained during the synapse-threshold synergistic learning process, which might provide the basis to SNNs for yielding higher accuracy.
    
    There is no doubt that the distribution of spike thresholds largely determines the firing rates of neurons in SNNs. The threshold distributions of the neurons exhibit notable differences in the two STL-SNNs trained with different initial conditions [see Figs.~\ref{Fig4}(a) and~\ref{Fig4}(b)]. Compared to the first STL-SNN, the spike thresholds of neurons in the convolutional layer are considerably higher in the second STL-SNN due to the larger initial threshold condition. For the same inputs, this result leads to lower firing rates for the convolutional layer in the second STL-SNN than in the first STL-SNN. Additional comparisons reveal that the neurons in the fully connected layer and the voting layer tend to maintain their firing rates by suppressing strong inputs [Fig.~\ref{Fig4}(a)] or amplifying weak inputs [Fig.~\ref{Fig4}(b)] from upstream neurons by modulating neural thresholds. Theoretically, this superior feature may not only provide an underlying mechanism for maintaining stable information propagation with appropriate firing rates but may also markedly reduce the sensitivity of SNNs to the initial spike thresholds.

	\subsubsection{Comparison with Equivalent Heterogeneous SNNs}

   	\begin{figure}[t]
		\centering
		\includegraphics[width=8.89 cm]{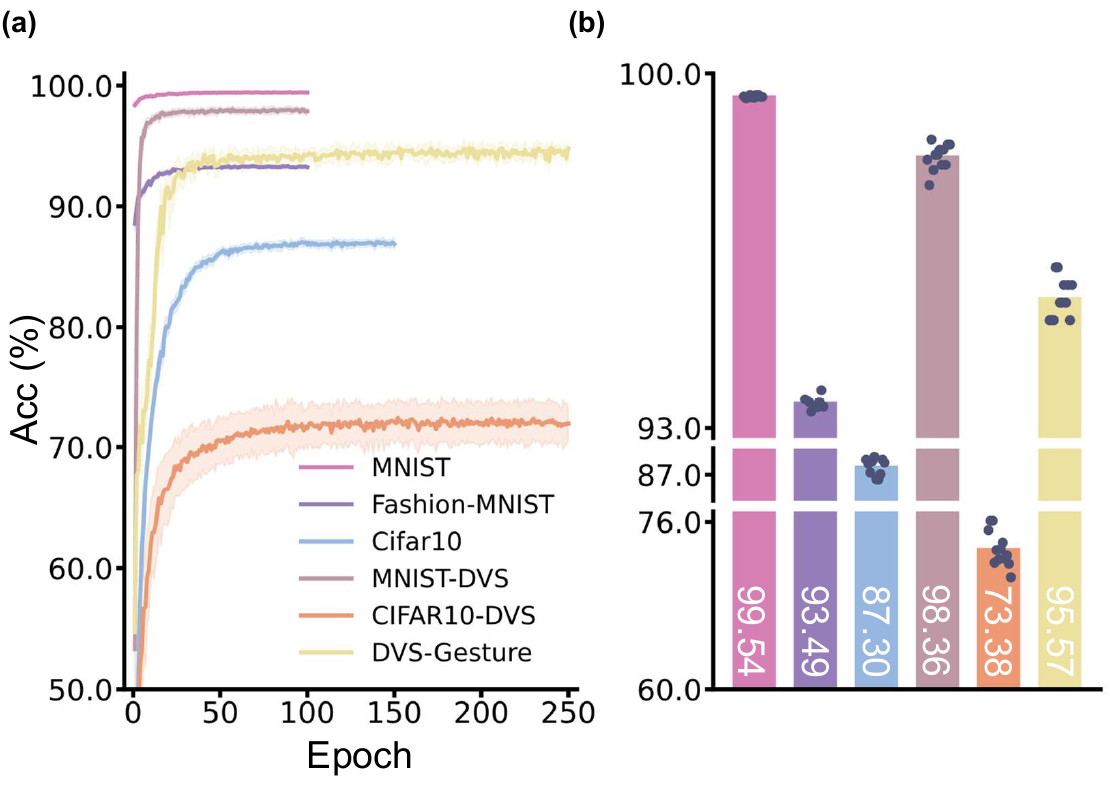} 
		\caption{The performance of Hete-SNNs on different static and neuromorphic datasets. (a) The average training curves of Hete-SNNs on different datasets. (b) The top-1 accuracy of each trial (dot) and the average top-1 accuracy (histogram). We repeat all experiments 10 trials.}
		\label{Fig5} 
	\end{figure}
    
    Neuronal heterogeneity is known to exist at different scales throughout the brain. Numerous studies have shown that neuronal heterogeneity plays several functional roles in neural computation~\cite{gjorgjieva2016computational} and can promote robust learning in SNNs~\cite{perez2021neural}. A question that naturally arises is whether the superiority of our proposed STL-SNN model can be completely attributed to the heterogeneity of spike thresholds. Here, we attempt to answer this question by establishing equivalent SNNs with threshold heterogeneity (Hete-SNNs). For each dataset, we first introduce an exactly same level of threshold heterogeneity into the Hete-SNN layer-by-layer,  by randomly shuffling the corresponding  spike thresholds in the best pre-trained STL-SNN. Then, we train such Hete-SNN with the synaptic learning approach. By employing equivalent Hete-SNNs, we can assess the impact of spike heterogeneity during synergistic learning in STL-SNNs.

    Fig.~\ref{Fig5} shows the training curves and accuracies of Hete-SNNs trained on different static and neuromorphic datasets. A more detailed comparison among the results of the Hete-SNNs and other types of SNNs is provided in Tab.~\ref{Tab2}. By comparing the Hete-SNN results with the SL-SNN results, we confirm that introducing the threshold heterogeneity into the Hete-SNNs improves the classification performance on most datasets (MNIST, Fashion-MNIST, MNIST-DVS and DVS-Gesture). This observation is consistent with previous findings reported in~\cite{perez2021neural}, implying that neuronal heterogeneity is an effective mechanism for improving the learning capability of SNNs. Interestingly, we highlight that the STL-SNNs significantly outperform the equivalent Hete-SNNs in terms of both the best top-1 accuracy and average top-1 accuracy on all datasets [see Tab.~\ref{Tab2}]. Our findings emphasize the importance of complementary effects between synaptic weights and spike thresholds that are naturally acquired during the synergistic learning process.

	\begin{figure}[t]
		\centering
		\includegraphics[width=8.89cm]{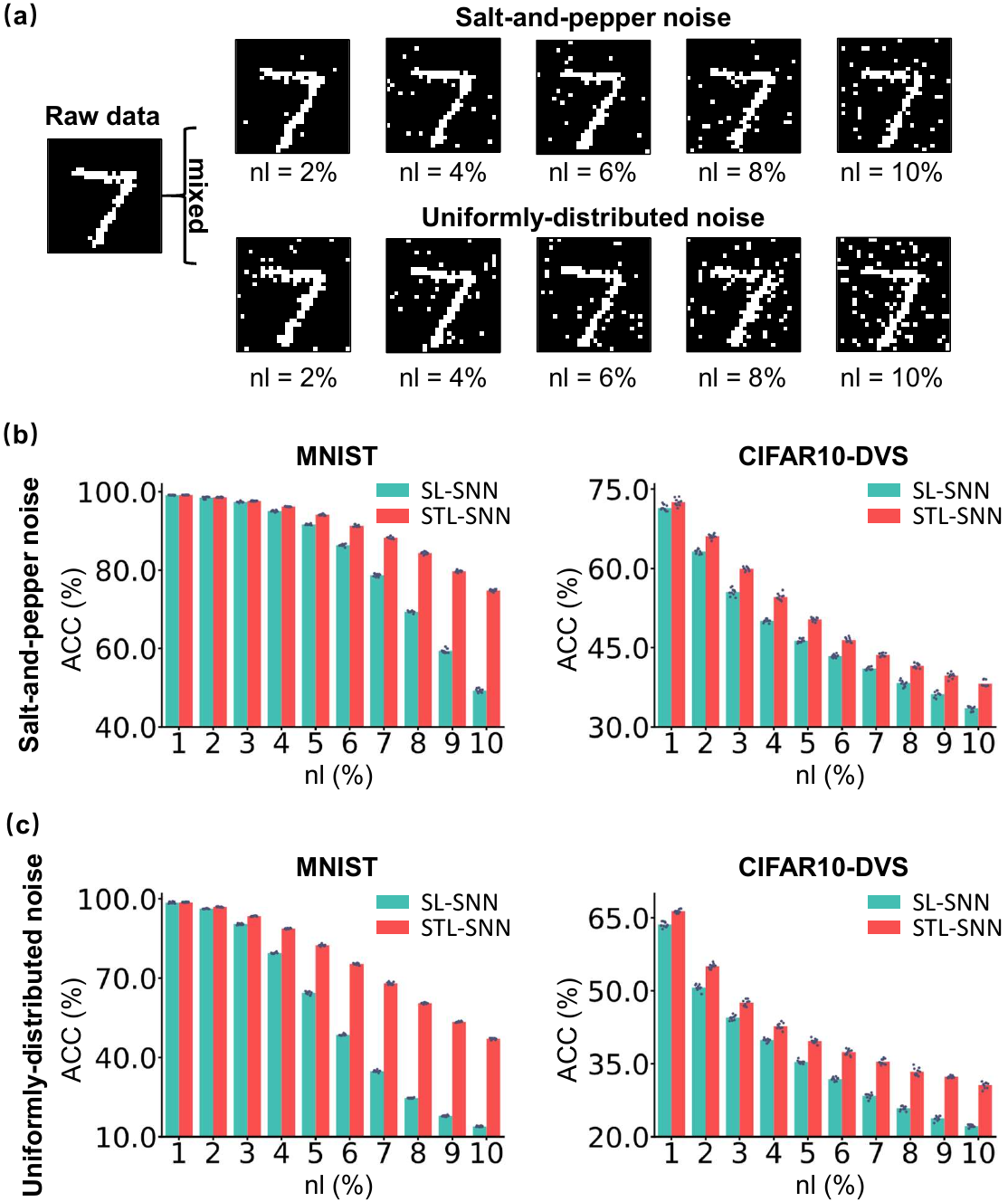} 
		\caption{Evaluation of fault tolerance to noise for both STL-SNNs and SL-SNNs. (a) An illustration of noise-mixed data with two different types of noise: salt-and-pepper noise and uniformly-distributed noise. The proportion of the noise region on each data is denoted as the noise level~(nl). A detailed comparison of STL-SNNs and SL-SNNs for different levels of salt-and-pepper noise (b) and uniformly-distributed noise (c). The top-1 accuracy of each trial (dot) and the average top-1 accuracy (histogram). We repeat all experiments 10 trials. Compared with the SL-SNN model, the STL-SNN model exhibits stronger robustness to noisy data.}
		\label{Fig6} 
	\end{figure}

    \subsubsection{Robustness to Noisy Data}    
 	We next examine the robustness of the SNNs trained by the proposed synergistic learning approach. A high fault tolerance to noisy data is critical for biologically-inspired SNNs, as this guarantees their robust learning capability in uncertain environments~\cite{johnson2017homeostatic}. To evaluate the robustness of the SNNs with respect to noise, we directly apply the pre-trained models on standard MNIST and CIFAR10-DVS datasets and evaluate the models using two types of noise-mixed data, namely, data mixed with salt-and-pepper noise and data mixed with uniformly distributed noise, as shown in Fig.~\ref{Fig6}(a). The results presented in Figs.~\ref{Fig6}(b) and 6(c) show the performance of STL-SNNs and SL-SNNs for different types of noise. Although the classification accuracies of both models decrease with increasing noise level, the STL-SNNs stably outperform the SL-SNNs. Furthermore, the superiority of the STL-SNNs becomes more apparent at higher noise levels [see Figs.~\ref{Fig6}(b) and (c)]. Overall, the above results provide compelling evidence that synapse-threshold synergistic learning can improve the robustness of the network to noise, thus potentially enhancing the fault tolerance of spiking neural networks.
    
	\subsubsection{Analysis of Stability and Efficiency} 
\begin{figure}[t]
   	\centering
   	\includegraphics[width=8.89 cm]{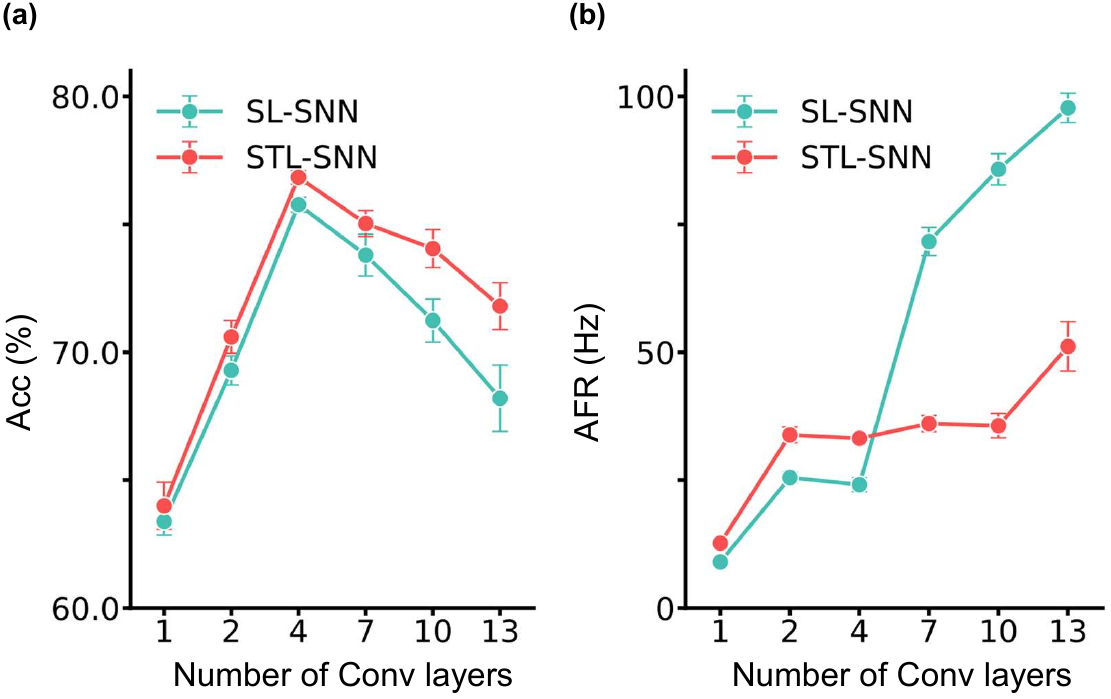} 
   	\caption{Stability and efficiency of SL-SNNs and STL-SNNs on the CIFAR10-DVS dataset. For a detailed comparison, we consider networks with different numbers of convolutional layers. The average top-1 accuracy (a) and average firing rate (AFR) over the whole network (b) versus the number of convolutional layers are plotted. Each result is represented as the mean and standard deviation over 10 trials. }
   	\label{Fig7} 
   \end{figure}

    To better understand the stability of the synapse-threshold synergistic learning model, we perform additional experiments on the complex CIFAR10-DVS dataset by using SNNs with different structures. For simplicity, we employ the same structure for the fully connected layers and voting layers as shown in Fig.~\ref{Fig1} (bottom), and we introduce convolutional layers with different depths to construct the networks. Fig.~\ref{Fig7}(a) summarizes the classification performance of the STL-SNNs and SL-SNNs with different convolutional layers in detail. In general, the STL-SNNs exhibit higher accuracies than the SL-SNNs with the same structures. In particular, this superior performance becomes more significant as the convolutional layers deepen. These findings verify the stability of our proposed synergistic learning approach for training SNNs.
    
    On the other hand, the power consumption in SNNs is believed to be mainly due to the generation of spike events. Thus, the average firing rates of neurons can serve as an essential index for measuring the efficiency of SNNs, and this metric has been widely used in previous studies~\cite{chen2022adaptive}. To comprehensively evaluate the efficiency of SNNs trained by different learning models, we perform additional experiments and compute the average firing rates of neurons in networks with different structures. Compared with the SL-SNN results, we observe that SNNs trained with the synapse-threshold synergistic learning approach can maintain relatively stable average firing rates that are neither too low nor too high [see Fig.~\ref{Fig7}(b)], regardless of the network structure. We postulate that this result occurs because of the threshold adjustment ability benefiting from the synergistic learning approach. As a consequence, this results in a slightly higher but acceptable energy consumption for STL-SNNs with shallower structures and a notably lower energy consumption for STL-SNNs with deeper network structures [Fig.~\ref{Fig7}(b)].
   
	\subsection{Joint Decision Framework for SNNs}
	As described above, we employ a commonly used rate-based decoding scheme for SNNs to distinguish different classification categories~\cite{wu2019direct}. However, because the spike activity is discrete, an inevitable decision difficulty may occasionally occur when more than two output classes share the highest spiking activity. Previous studies have suggested that this decision difficulty can be alleviated by using a longer time window or larger population size~\cite{cheng2020lisnn, fang2021incorporating}. Inspired by the concept of ensemble learning~\cite{rokach2010ensemble}, we provide another simple strategy based on the joint voting of independent networks to effectively reduce potential decision difficulties in SNNs with the rate-based decoding scheme.

	\begin{figure}[t]
	\centering
	\includegraphics[width=8.89 cm]{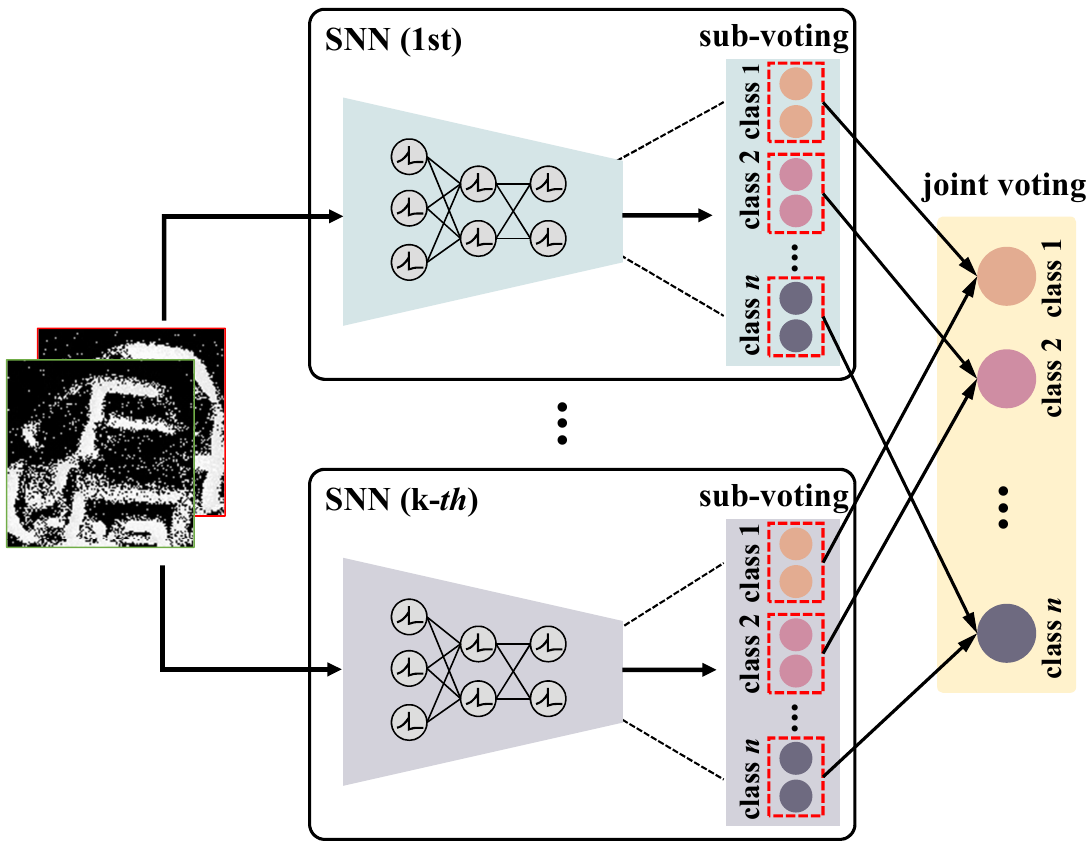} 
	\caption{A joint decision framework (JDF) for SNNs with rate-based decoding schemes. The JDF consists of $k$ independent SNNs that are trained with different initial conditions in the synaptic weights. A joint voting layer is constructed by counting spiking events of neurons in the sub-voting classes from all independent SNNs.}
	\label{Fig8} 
    \end{figure} 

    \begin{figure*}[t]
	\centering
	\includegraphics[width=18.1cm]{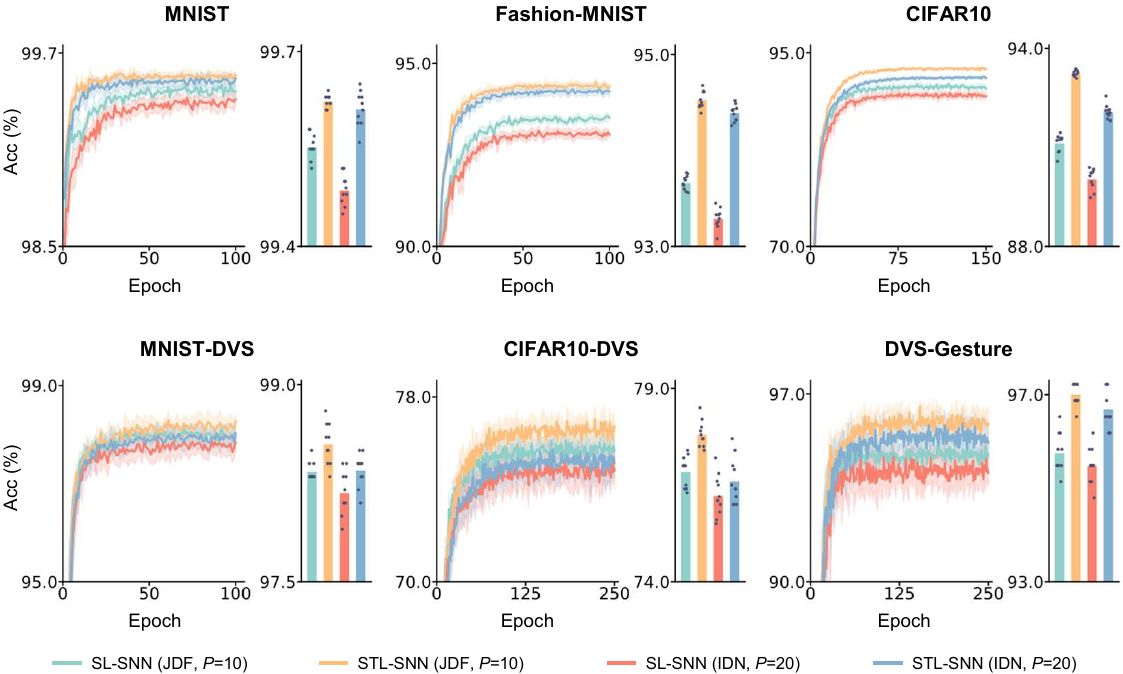} 
	\caption{Performance of SL-SNNs and STL-SNNs under the joint decision framework composed of two ($k=2$) independent SNNs with a population size of $P=10$ for each class in the sub-voting layer. On each dataset, we employ the same network structure as that given in Fig.~\ref{Fig1} (bottom) to construct the joint decision network. To evaluate the effect of joint decision strategy, independent (IND) STL-SNNs and SL-SNNs are also established by doubling the population size ($P=20$) for each class in the voting layers. For each dataset, the average training curves with the standard deviations are presented for different types of SNNs (left), and both the top-1 accuracy of each trial (dot) and the average top-1 accuracy (histogram) are also plotted (right) for a detailed comparison. All experiments are repeated for 10 trials. The results show that the joint decision can significantly improve the performance of SNNs.}
	\label{Fig9} 
     \end{figure*}      
   
    In Fig.~\ref{Fig8}, we schematically show the joint decision framework~(JDF) for SNNs with the rate-based decoding scheme. In brief, the proposed JDF structure is composed of multiple SNNs, and each SNN is trained independently with different initial synaptic weights. During the inference phase on the test dataset, we collect the spiking events of neurons in each sub-voting class from multiple SNNs and determine the joint voting decision based on the highest level of spiking activity to predict the classification result. In principle, SNNs trained with different initial conditions are likely to enrich the output diversity, which may play a potential role in relieving the possible decision difficulties via the JDF.
    
	Fig.~\ref{Fig9} depicts the results of SNNs trained by the synaptic learning approach and synapse-threshold synergistic learning approach under the JDF ($k=2$). By comparing the best top-1 accuracy and average top-1 accuracy [see Tab.~\ref{Tab2}], we observe that the joint decision of two independent networks can significantly enhance the classification performance for different types of SNNs on all static and neuromorphic datasets. However, a question that may naturally arise is: whether the capability of the joint decision strategy is simply a consequence of increasing the number of neurons in the voting layer. To examine this possibility, we further construct independent SNNs for different learning approaches with equivalent numbers of neurons in the voting layers as those in the joint decision networks, and the results of these SNNs ($P=20$) are also presented in Fig.~\ref{Fig9}. Compared to the approach of simply increasing the population size of each class in the voting layer, the joint decision strategy exhibits a much stronger capability to deal with possible decision difficulties in SNNs caused by the rate-based decoding scheme.  These evidence indicates that the JDF might be a generalized strategy to improve the performance of SNNs. However, such strategy increases the number of model parameters, thus leading to higher computational complexity.
	
	Additionally, we also demonstrate that the STL-SNNs~(JDF) can achieve better performance than the SL-SNNs~(JDF) on different benchmark datasets [Fig.~\ref{Fig9} and Tab.~\ref{Tab2}]. This result is consistent with our above observations and further highlights the universality of appropriate synergies between synaptic weights and spike thresholds for training high-performance SNNs.

	\subsection{Validation on Challenging Datasets and Complicated Tasks}	
\begin{figure}[t]
		\centering
		\includegraphics[width=8.89 cm]{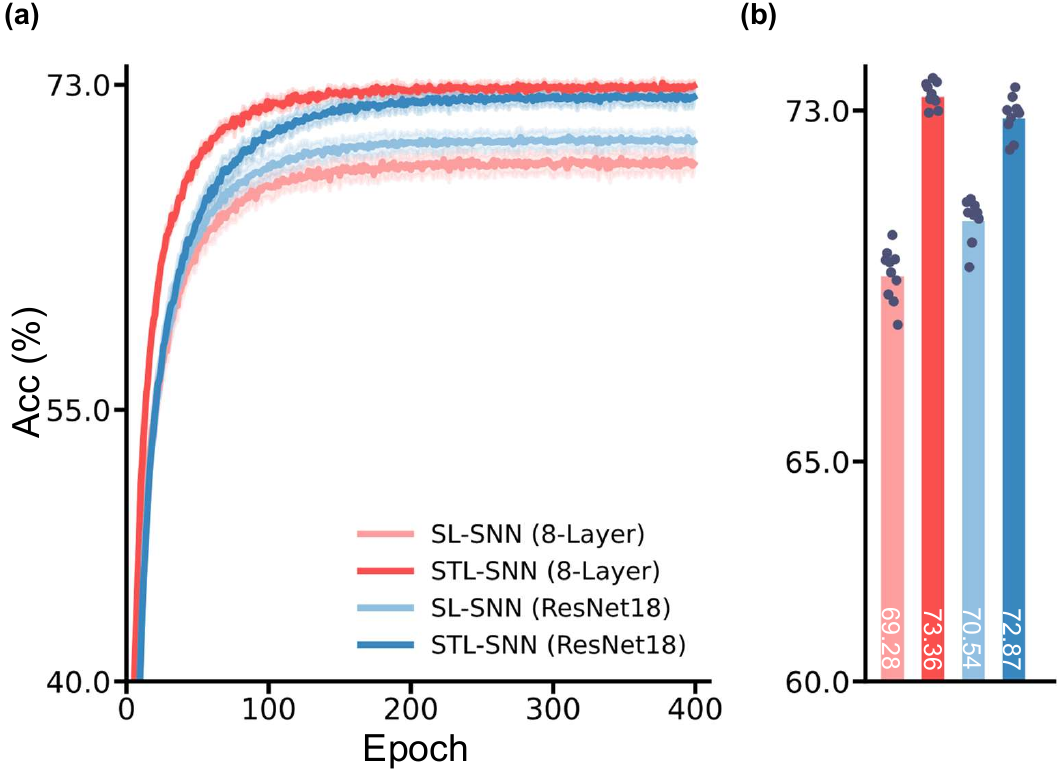} 
		\caption{Performance of SL-SNNs and STL-SNNs on the CIFAR100 dataset. (a) The average training curves with standard deviations for different SNN models. (b) The top-1 accuracy of each trial (dot) and the average top-1 accuracy (histogram). Two types of network structures (the 8-Layer SNN structure and the ResNet18 structure) are used in this study. All experiments are repeated for 10 trails. For both types of network structures, the STL-SNN models maintain their advantages over the SL-SNN models.}
		\label{Fig10} 
	\end{figure}

	To further validate the effectiveness of our proposed method, we conduct additional experiments on the more challenging CIFAR100 dataset. Here, we consider two different types of network structures, including the 8-Layer SNN given in Fig.~\ref{Fig2} and the ResNet18 SNN, which is widely used to evaluate learning algorithms on CIFAR100. Similar to a previous work~\cite{li2021differentiable}, the ResNet18 SNN used in our study utilizes traditional fully connected layer as the decoder and calculates the loss function using cross-entropy. The detailed training hyperparameters for both network structures are given in Tab.~\ref{Tab1}. As expected, we find that the SNNs trained with the synapse-threshold synergistic learning method consistently outperform the baseline SNNs for the 8-Layer SNN  (best: 73.74\%; average: 73.36$\pm$0.26\%) and the ResNet18 SNN  (best: 73.53\%; average: 72.87$\pm$0.42\%) [Fig.~\ref{Fig10} and Tab.~\ref{Tab4}]. Remarkably, the performance of our STL-SNN models can be further improved by a large margin (more than 2\%) under the JDF framework ($k=2$). Additionally, we compare the performance of STL-SNN models with that of other existing state-of-the-art works in Tab.~\ref{Tab4}. It should be noted that given the challenging nature of the CIFAR100 dataset, the SNN models developed in previous studies often rely on hybrid learning approaches or ANN-to-SNN conversions to achieve better results. Our results demonstrate that the proposed STL-SNN models can achieve competitive performance with existing state-of-the-art models by using a simple spike-based BP method with a limited number of time steps [Tab.~\ref{Tab4}].

\begin{figure}[t]
		\centering
		\includegraphics[width=8.89 cm]{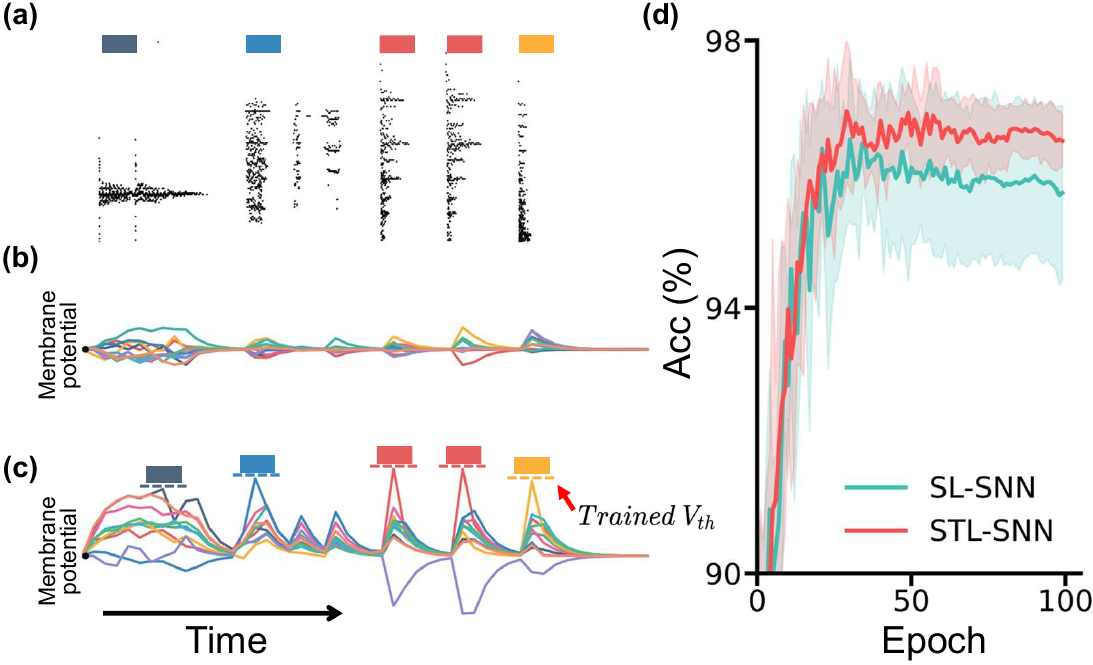} 
		\caption{Performance of SL-SNN and STL-SNN models for the unsegmented sensory event detection task. (a) The spike pattern of an unsegmented sensory stream that contains 5 individual events. The colored rectangles indicate different event classes. (b-c) Typical examples of the membrane potential traces of output neurons in the STL-SNN model. Different colors represent distinct classes. The desired output neuron cannot respond to its corresponding events before training (b), whereas after training it fires a spike when a corresponding event occurs. (d) The average training curves with standard deviations for the SL-SNN and STL-SNN models, and both curves are averaged over 10 trails.}
		\label{Fig11} 
	\end{figure}

Moreover, the temporal credit assignment (TCA) is another challenging task that aims to discover the predictive features hidden in distracting background streams with delayed feedback. As a typical TCA task, unsegmented sensory event detection in traditional solutions requires heavy labor or complex algorithms~\cite{gu2019stca}. To demonstrate the capability of our proposed STL-SNN model in processing TCA tasks, we employ a subset of the RWCP dataset~\cite{nakamura2000acoustical, gu2019stca}, which includes 10 classes of isolated sound event samples from the original dataset and constructs unsegmented sound event streams by randomly splicing five individual events [Fig.~\ref{Fig11}(a)]. For a fair comparison, we adopt the same architecture as that used in a previous work~\cite{gu2019stca} [see Fig.\ref{Fig2}], and the detailed values of the hyperparameters of different SNN models are listed in Tab.1. By tracking the membrane potential traces of different output neurons [see Figs.~\ref{Fig11}(a)-(c)], we observe that the trained desired neuron in the STL-SNN model can emit a spike determined by its own threshold within a short time window following a sound event. Otherwise, the desired neuron tends to remain below the trained threshold. In Fig.~\ref{Fig11}(d), we plot the average training curves with standard deviations for both the SL-SNN and STL-SNN models. Clearly, our results suggest that the proposed synapse-threshold synergistic learning method can also greatly improve the performance of SNNs on the TCA tasks.

	\section{Discussion and Conclusion} 
	Recently, spiking neural networks have attracted extensive attention in the field of artificial intelligence because of their low-power computation and excellent performance in many realistic intelligent tasks~\cite{ghosh2009spiking,tavanaei2019deep, 8720035}. However, the direct training of SNNs is difficult, and the development of highly efficient learning methods for SNNs remains a remarkable challenge. In this work, we proposed a novel synapse-threshold synergistic learning approach for SNNs that incorporates the spike threshold as a learnable parameter. By performing a systematic evaluation, we showed that the proposed synergistic learning approach can be used to train SNNs on different static and neuromorphic datasets, and the resulting performance is significantly higher than that of SNNs trained with single-learning methods. In addition, we introduced a joint decision framework for SNNs with the rate-based decoding scheme and demonstrated that joint voting of independent networks can effectively prevent possible decision difficulties.

\begin{table*}[t]
	\centering
	\caption{\label{Tab4} Comparison with existing state-of-the-art results on the CIFAR100 dataset.}
	\renewcommand\arraystretch{1.2}
	\begin{tabular}{lcccc}
		\hline
		Model              & Architecture   & Method     & T     & Accuracy      \\ \hline
		Diet-SNN~\cite{rathi2021diet}       & ResNet20    & ANN2SNN+Spike-based BP  & 5   & 64.07\%     \\
		Diet-SNN~\cite{rathi2021diet}       & VGG16       & ANN2SNN+Spike-based BP  & 5   & 69.67\%       \\
		Dspike~\cite{li2021differentiable}  & ResNet18    & ANN2SNN+Spike-based BP  & 4   & 73.35$\pm$0.14\%\\
		SNN~\cite{deng2021comprehensive}    & 11-Layer SNN   & Spike-based BP          & 8   & 57.83\%       \\
		EDB~\cite{zhu2022training}          & VGG11       & Spike-based BP          & 16  & 63.97\%       \\
		IM-ESG~\cite{guoloss}               & VGG16       & Spike-based BP          & 5   & 70.18$\pm$0.09\%\\
		IC-SNN~\cite{li2022ic}              & ResNet20    & ANN2SNN                 & 256 & 65.60\%       \\
		SNM-NeuronNorm~\cite{wang2022signed}& VGG16       & ANN2SNN                 & 32  & 71.80\%       \\
		SNM-NeuronNorm~\cite{wang2022signed}& ResNet18    & ANN2SNN                 & 32  & 74.48\%       \\
		\hline
		STL-SNN~(Best top-1)                & 8-Layer SNN    & Spike-based BP          & 8   & 73.74\%       \\
		STL-SNN~(Average top-1)             & 8-Layer SNN    & Spike-based BP          & 8   & 73.36$\pm$0.26\%\\
		STL-SNN~(JDF)~(Best top-1)          & 8-Layer SNN    & Spike-based BP          & 8   & 76.00\%       \\
		STL-SNN~(JDF)~(Average top-1)       & 8-Layer SNN    & Spike-based BP          & 8   & 75.80$\pm$0.16\%\\
		\hline
		STL-SNN~(Best top-1)                & ResNet18    & Spike-based BP          & 4   & 73.53\%       \\
		STL-SNN~(Average top-1)             & ResNet18    & Spike-based BP          & 4   & 72.87$\pm$0.42\%\\
		STL-SNN~(JDF)~(Best top-1)          & ResNet18    & Spike-based BP          & 4   & 76.15\%       \\
		STL-SNN~(JDF)~(Average top-1)       & ResNet18    & Spike-based BP          & 4   & 75.65$\pm$0.26\%\\
		\hline
				 
	\end{tabular}
\end{table*}

	Hybrid learning is a widely used strategy that combines different training methods to improve the performance of artificial intelligence models~\cite{wu2022brain, zhang2021self}. In essential, the synapse-threshold synergistic learning approach developed in this study is a typical hybrid learning model. Indeed, synaptic plasticity and intrinsic non-synaptic plasticity both contribute to biological learning in the brain~\cite{schrauwen2008improving, stemmler1999voltage, joshi2009rules}. A natural question to ask is whether the learning capability of SNNs can be improved by integrating these two types of learning mechanisms. To the best of our knowledge, however, most existing hybrid learning models for SNNs have been established based on the concept of synaptic plasticity, and only limited studies have considered the use of non-synaptic learning mechanisms to train SNNs for intelligent tasks~\cite{huang2016adaptive, zhang2020intrinsic, zhang2019fast}. As a critical intrinsic property of neurons~\cite{fontaine2014spike, azouz2000dynamic}, the spike threshold is an idealized regulation target to be adjusted in SNNs. Unlike several previous approaches that try to update threshold-related parameters adaptively~\cite{li2012spike, salaj2021spike}, we treated the spike threshold as a learnable parameter and derived a direct synapse-threshold synergistic training method for SNNs based on a supervised learning approach. Our results demonstrated that biologically plausible synergies between synaptic weights and spike thresholds can markedly elevate the accuracy of SNNs. Interestingly, a recent study also showed that simultaneously learning synaptic weights and membrane time constants in SNNs can promote network performance on different classification tasks~\cite{fang2021incorporating}. This result is not surprising because, similar to the spike threshold, the time constant of the membrane potential is another learnable parameter that modulates the firing rate of neurons; however, these parameters may have different impacts on neuronal dynamics. Overall, these preliminary results indicate that the capability of SNNs can be improved by suitably synergizing synaptic and non-synaptic learning mechanisms.

	Moreover, a better understanding of biologically-inspired intrinsic non-synaptic plasticity may provide insights into the development of hybrid synergistic learning models for SNNs. Recent experiential studies have revealed a variety of underlying non-synaptic learning mechanisms that have been shown to affect the intrinsic excitability of neurons at the cellular level~\cite{zhang2003other, mozzachiodi2010more}. Furthermore, different types of non-synaptic plasticity have been reported to influence the electrical properties of neurons from various perspectives, including but not limited to synaptic integration~\cite{nickel2010evolutionary}, subthreshold neuronal dynamics~\cite{rubinov2011neurobiologically} and spike generation~\cite{azouz2003adaptive}. Accordingly, it is reasonable to postulate that the performance and training speed of SNNs can both be improved by combining complementary non-synaptic learning mechanisms with synaptic plasticity, and this prediction should be tested in future studies.

	It has been known that neuronal heterogeneity may arise from different biophysical substrates and can be considered as an efficient strategy for the brain~\cite{koch1999complexity}. In particular, previous studies have suggested that heterogeneity in the brain may play a functional role in neural computation and cognitive processes, such as efficient coding~\cite{marsat2010neural}, working memory~\cite{kilpatrick2013optimizing} and robust learning~\cite{perez2021neural}. Our results presented here also demonstrated that spiking neural networks with appropriate levels of threshold heterogeneity achieve higher accuracies than homogeneous networks for most tasks, but synergistic training between synaptic weights and spike thresholds can further improve their performance. Importantly, we found that the superiority of synapse-threshold synergistic learning is especially noticeable for several difficult tasks with complicated scenarios, such as tasks on the CIFAR10 and CIFAR10-DVS datasets [see Tab.~\ref{Tab2}]. As a consequence, we hypothesize that the seemingly random spike thresholds observed in experiments can serve as an important component of biological learning, which may be at least acquired partly during the learning process in the brain.

	Theoretically, high fault tolerance to noisy data and low power consumption during computations are both essential factors for real-time information processing on neuromorphic chips~\cite{spyrou2021neuron}. Our analysis revealed that SNNs trained by the synergistic learning approach exhibit strong robustness to various types of noise. This advantageous feature provides the established STL-SNNs with a strong robust learning capability, reducing the influence of unavoidable noise generated by complicated external environments and hardware restrictions. On the other hand, benefiting from the appropriate synergies between synaptic weights and spike thresholds, we further showed that STL-SNNs with different depths can achieve excellent performance while maintaining suitable firing rates that are neither too low nor too high. This superior feature has been demonstrated to guarantee reasonable power consumption for STL-SNNs and, in particular, may economize the energy requirements of networks with deep structures. Thus, the advantages of our proposed synergistic learning approach may be helpful for designing highly efficient online SNN models on neuromorphic chips.
	
	It is worth noting that the learnable thresholds have also been employed in several previous studies~\cite{pellegrini2021low, meng2022training, jiang2021few, rathi2021diet}. However, most of the SNNs  developed in these studies share the same threshold for neurons in the same layer~\cite{meng2022training, jiang2021few, rathi2021diet}. In contrast, our proposed synergistic learning strategy is different from these prior works and aims to simultaneously train the synaptic weights and thresholds of neurons at the single-neuron level. Indeed, our design philosophy is more biologically plausible, as the spike threshold is an important intrinsic property of biological neurons, and different neurons have different thresholds in the brain regardless of their type or origin. As discussed above, our findings emphasize the importance of complementary effects between synaptic weights and spike thresholds that are naturally acquired during the synergistic learning process. Theoretically, such complementary effects tend to occur at the single-neuron level and might be weakened if the shared learnable thresholds are introduced into SNNs. On the other hand, although the study in~\cite{pellegrini2021low} also used trainable thresholds in SNNs at the single-neuron level, their experimental results imply that learning thresholds are not needed for SNNs, which contradicts the findings in our work and other well-established studies~\cite{meng2022training, jiang2021few, rathi2021diet}.
		
    To summarize, we developed a synapse-threshold synergistic learning approach for directly training SNNs. We have systematically verified that this synergistic learning can not only equip SNNs with competitive performance on various datasets with different  tasks, but also endow them with strong robustness, stable signal transmission and reasonable energy consumption. These findings emphasize that appropriately incorporating biologically-inspired learning mechanism into the training of SNNs can significantly promote their performance, which may provide a promising approach for optimizing existing SNN learning methods.

    \section*{Acknowledgments} We sincerely thank Shikuang Deng, Runhao Jiang and Pengjie Gu for valuable discussions during the revision of this study.

	\bibliography{IEEEabrv,ref}

\begin{thebibliography}{10}
\providecommand{\url}[1]{#1}
\csname url@samestyle\endcsname
\providecommand{\newblock}{\relax}
\providecommand{\bibinfo}[2]{#2}
\providecommand{\BIBentrySTDinterwordspacing}{\spaceskip=0pt\relax}
\providecommand{\BIBentryALTinterwordstretchfactor}{4}
\providecommand{\BIBentryALTinterwordspacing}{\spaceskip=\fontdimen2\font plus
\BIBentryALTinterwordstretchfactor\fontdimen3\font minus
  \fontdimen4\font\relax}
\providecommand{\BIBforeignlanguage}[2]{{%
\expandafter\ifx\csname l@#1\endcsname\relax
\typeout{** WARNING: IEEEtran.bst: No hyphenation pattern has been}%
\typeout{** loaded for the language `#1'. Using the pattern for}%
\typeout{** the default language instead.}%
\else
\language=\csname l@#1\endcsname
\fi
#2}}
\providecommand{\BIBdecl}{\relax}
\BIBdecl

\bibitem{koch1999complexity}
C.~Koch and G.~Laurent, ``Complexity and the nervous system,'' \emph{Science},
  vol. 284, no. 5411, pp. 96--98, 1999.

\bibitem{daoudal2003long}
G.~Daoudal and D.~Debanne, ``Long-term plasticity of intrinsic excitability:
  learning rules and mechanisms,'' \emph{Learn. Mem.}, vol.~10, no.~6, pp.
  456--465, 2003.

\bibitem{gosak2022networks}
M.~Gosak, M.~Milojevi{\'c}, M.~Duh, K.~Skok, and M.~Perc, ``Networks behind the
  morphology and structural design of living systems,'' \emph{Phys. Life Rev.},
  vol.~41, pp. 1--21, 2022.

\bibitem{klinshov2022rate}
V.~V. Klinshov, A.~V. Kovalchuk, I.~Franovi{\'c}, M.~Perc, and M.~Svetec,
  ``Rate chaos and memory lifetime in spiking neural networks,'' \emph{Chaos
  Solitons Fractals}, vol. 158, p. 112011, 2022.

\bibitem{li2023brain}
G.~Li, L.~Deng, H.~Tang, G.~Pan, Y.~Tian, K.~Roy, and W.~Maass, ``Brain
  inspired computing: A systematic survey and future trends,'' \emph{TechRxiv},
  2023, {\color{blue}{doi:10.36227/techrxiv.21837027.v1}}.

\bibitem{maass1997networks}
W.~Maass, ``Networks of spiking neurons: the third generation of neural network
  models,'' \emph{Neural Netw.}, vol.~10, no.~9, pp. 1659--1671, 1997.

\bibitem{pei2019towards}
J.~Pei, L.~Deng, S.~Song, M.~Zhao, Y.~Zhang, S.~Wu, G.~Wang, Z.~Zou, Z.~Wu,
  W.~He \emph{et~al.}, ``Towards artificial general intelligence with hybrid
  tianjic chip architecture,'' \emph{Nature}, vol. 572, no. 7767, pp. 106--111,
  2019.

\bibitem{9665763}
A.~Zhang, Y.~Han, Y.~Niu, Y.~Gao, Z.~Chen, and K.~Zhao, ``Self-evolutionary
  neuron model for fast-response spiking neural networks,'' \emph{IEEE Trans.
  Cogn. Dev. Syst.}, vol.~14, no.~4, pp. 1766--1777, 2022.

\bibitem{8981937}
Y.~Wang, Y.~Xu, R.~Yan, and H.~Tang, ``Deep spiking neural networks with binary
  weights for object recognition,'' \emph{IEEE Trans. Cogn. Dev. Syst.},
  vol.~13, no.~3, pp. 514--523, 2021.

\bibitem{8720035}
L.~Cheng, Y.~Liu, Z.-G. Hou, M.~Tan, D.~Du, and M.~Fei, ``A rapid spiking
  neural network approach with an application on hand gesture recognition,''
  \emph{IEEE Trans. Cogn. Dev. Syst.}, vol.~13, no.~1, pp. 151--161, 2021.

\bibitem{sigaki2020learning}
H.~Y. Sigaki, E.~K. Lenzi, R.~S. Zola, M.~Perc, and H.~V. Ribeiro, ``Learning
  physical properties of liquid crystals with deep convolutional neural
  networks,'' \emph{Sci. Rep.}, vol.~10, no. 7664, pp. 1--10, 2020.

\bibitem{li2012spike}
C.~Li and Y.~Li, ``A spike-based model of neuronal intrinsic plasticity,''
  \emph{{IEEE} Trans. Auton. Mental Develop.}, vol.~5, no.~1, pp. 62--73, 2013.

\bibitem{liu2020effective}
Q.~Liu, H.~Ruan, D.~Xing, H.~Tang, and G.~Pan, ``Effective {AER} object
  classification using segmented probability-maximization learning in spiking
  neural networks,'' in \emph{Proc. AAAI Conf. Artif. Intell.}, 2020, pp.
  1308--1315.

\bibitem{diehl2015unsupervised}
P.~U. Diehl and M.~Cook, ``Unsupervised learning of digit recognition using
  spike-timing-dependent plasticity,'' \emph{Front. Comput. Neurosci.}, vol.~9,
  p.~99, 2015.

\bibitem{8354825}
C.~Lee, G.~Srinivasan, P.~Panda, and K.~Roy, ``Deep spiking convolutional
  neural network trained with unsupervised spike-timing-dependent plasticity,''
  \emph{IEEE Trans. Cogn. Dev. Syst.}, vol.~11, no.~3, pp. 384--394, 2019.

\bibitem{wu2021tandem}
J.~Wu, Y.~Chua, M.~Zhang, G.~Li, H.~Li, and K.~C. Tan, ``A tandem learning rule
  for effective training and rapid inference of deep spiking neural networks,''
  \emph{IEEE Trans. Neural Netw. Learn. Syst.}, vol.~34, no.~1, pp. 446--460,
  2023.

\bibitem{diehl2015fast}
P.~U. Diehl, D.~Neil, J.~Binas, M.~Cook, S.-C. Liu, and M.~Pfeiffer,
  ``Fast-classifying, high-accuracy spiking deep networks through weight and
  threshold balancing,'' in \emph{Proc. Int. Joint Conf. Neural Netw.}, 2015.

\bibitem{wu2018spatio}
Y.~Wu, L.~Deng, G.~Li, J.~Zhu, and L.~Shi, ``Spatio-temporal backpropagation
  for training high-performance spiking neural networks,'' \emph{Front.
  Neurosci.}, vol.~12, p. 331, 2018.

\bibitem{wu2019direct}
Y.~Wu, L.~Deng, G.~Li, J.~Zhu, Y.~Xie, and L.~Shi, ``Direct training for
  spiking neural networks: faster, larger, better,'' in \emph{Proc. AAAI Conf.
  Artif. Intell.}, vol.~33, no.~1, 2019, pp. 1311--1318.

\bibitem{cramer2022surrogate}
B.~Cramer, S.~Billaudelle, S.~Kanya, A.~Leibfried, A.~Gr{\"u}bl, V.~Karasenko,
  C.~Pehle, K.~Schreiber, Y.~Stradmann, J.~Weis \emph{et~al.}, ``Surrogate
  gradients for analog neuromorphic computing,'' \emph{Proc. Natl. Acad. Sci.
  U. S. A.}, vol. 119, no.~4, p. e2109194119, 2022.

\bibitem{mozafari2018first}
M.~Mozafari, S.~R. Kheradpisheh, T.~Masquelier, A.~Nowzari-Dalini, and
  M.~Ganjtabesh, ``First-spike-based visual categorization using
  reward-modulated {STDP},'' \emph{IEEE Trans. Neural Netw. Learn. Syst.},
  vol.~29, no.~12, pp. 6178--6190, 2018.

\bibitem{nicola2017supervised}
W.~Nicola and C.~Clopath, ``Supervised learning in spiking neural networks with
  {FORCE} training,'' \emph{Nat. Commun.}, vol.~8, no. 2208, pp. 1--15, 2017.

\bibitem{hao2020biologically}
Y.~Hao, X.~Huang, M.~Dong, and B.~Xu, ``A biologically plausible supervised
  learning method for spiking neural networks using the symmetric {STDP}
  rule,'' \emph{Neural Netw.}, vol. 121, pp. 387--395, 2020.

\bibitem{9950361}
Y.~Lin, Y.~Hu, S.~Ma, D.~Yu, and G.~Li, ``Rethinking pretraining as a bridge
  from {ANN}s to {SNN}s,'' \emph{IEEE Trans. Neural Netw. Learn. Syst.}, in
  press, {\color{blue}{doi:10.1109/TNNLS.2022.3217796}}.

\bibitem{wu2022brain}
Y.~Wu, R.~Zhao, J.~Zhu, F.~Chen, M.~Xu, G.~Li, S.~Song, L.~Deng, G.~Wang,
  H.~Zheng \emph{et~al.}, ``Brain-inspired global-local learning incorporated
  with neuromorphic computing,'' \emph{Nat. Commun.}, vol.~13, no.~65, pp.
  1--14, 2022.

\bibitem{zhang2021self}
T.~Zhang, X.~Cheng, S.~Jia, M.~ming Poo, Y.~Zeng, and B.~Xu,
  ``Self-backpropagation of synaptic modifications elevates the efficiency of
  spiking and artificial neural networks,'' \emph{Sci. Adv.}, vol.~7, no.~43,
  p. eabh0146, 2021.

\bibitem{legenstein2008learning}
R.~Legenstein, D.~Pecevski, and W.~Maass, ``A learning theory for
  reward-modulated spike-timing-dependent plasticity with application to
  biofeedback,'' \emph{PLoS Comput. Biol.}, vol.~4, no.~10, p. e1000180, 2008.

\bibitem{zeng2019short}
G.~Zeng, X.~Huang, T.~Jiang, and S.~Yu, ``Short-term synaptic plasticity
  expands the operational range of long-term synaptic changes in neural
  networks,'' \emph{Neural Netw.}, vol. 118, pp. 140--147, 2019.

\bibitem{zhang2003other}
W.~Zhang and D.~J. Linden, ``The other side of the engram: experience-driven
  changes in neuronal intrinsic excitability,'' \emph{Nat. Rev. Neurosci.},
  vol.~4, no.~11, pp. 885--900, 2003.

\bibitem{mozzachiodi2010more}
R.~Mozzachiodi and J.~H. Byrne, ``More than synaptic plasticity: role of
  nonsynaptic plasticity in learning and memory,'' \emph{Trends Neurosci.},
  vol.~33, no.~1, pp. 17--26, 2010.

\bibitem{schrauwen2008improving}
B.~Schrauwen, M.~Wardermann, D.~Verstraeten, J.~J. Steil, and D.~Stroobandt,
  ``Improving reservoirs using intrinsic plasticity,'' \emph{Neurocomputing},
  vol.~71, no. 7-9, pp. 1159--1171, 2008.

\bibitem{stemmler1999voltage}
M.~Stemmler and C.~Koch, ``How voltage-dependent conductances can adapt to
  maximize the information encoded by neuronal firing rate,'' \emph{Nat.
  Neurosci.}, vol.~2, no.~6, pp. 521--527, 1999.

\bibitem{joshi2009rules}
P.~Joshi and J.~Triesch, ``Rules for information maximization in spiking
  neurons using intrinsic plasticity,'' in \emph{Proc. Int. Joint Conf. Neural
  Netw.}, 2009, pp. 1456--1461.

\bibitem{azouz1999cellular}
R.~Azouz and C.~M. Gray, ``Cellular mechanisms contributing to response
  variability of cortical neurons in vivo,'' \emph{J. Neurosci.}, vol.~19,
  no.~6, pp. 2209--2223, 1999.

\bibitem{farries2010dynamic}
M.~A. Farries, H.~Kita, and C.~J. Wilson, ``Dynamic spike threshold and zero
  membrane slope conductance shape the response of subthalamic neurons to
  cortical input,'' \emph{J. Neurosci.}, vol.~30, no. 39, pp. 13180-13191,
  2010.

\bibitem{fontaine2014spike}
B.~Fontaine, J.~L. Pe{\~n}a, and R.~Brette, ``Spike-threshold adaptation
  predicted by membrane potential dynamics in vivo,'' \emph{PLoS Comput.
  Biol.}, vol.~10, no.~4, p. e1003560, 2014.

\bibitem{azouz2000dynamic}
R.~Azouz and C.~M. Gray, ``Dynamic spike threshold reveals a mechanism for
  synaptic coincidence detection in cortical neurons in vivo,'' \emph{Proc.
  Natl. Acad. Sci. U. S. A.}, vol.~97, no.~14, pp. 8110--8115, 2000.

\bibitem{huang2016adaptive}
C.~Huang, A.~Resnik, T.~Celikel, and B.~Englitz, ``Adaptive spike threshold
  enables robust and temporally precise neuronal encoding,'' \emph{PLoS Comput.
  Biol.}, vol.~12, no.~6, p. e1004984, 2016.

\bibitem{salaj2021spike}
D.~Salaj, A.~Subramoney, C.~Kraisnikovic, G.~Bellec, R.~Legenstein, and
  W.~Maass, ``Spike frequency adaptation supports network computations on
  temporally dispersed information,'' \emph{eLife}, vol.~10, p. e65459, 2021.

\bibitem{zhang2019information}
W.~Zhang and P.~Li, ``Information-theoretic intrinsic plasticity for online
  unsupervised learning in spiking neural networks,'' \emph{Front. Neurosci.},
  vol.~13, p.~31, 2019.

\bibitem{ding2021optimal}
J.~Ding, Z.~Yu, Y.~Tian, and T.~Huang, ``Optimal {ANN-SNN} conversion for fast
  and accurate inference in deep spiking neural networks,'' in \emph{Proc. 30th
  Int. Joint Conf. Artif. Intell.}, 2021, pp. 2328--2336.

\bibitem{sengupta2019going}
A.~Sengupta, Y.~Ye, R.~Wang, C.~Liu, and K.~Roy, ``Going deeper in spiking
  neural networks: {VGG} and residual architectures,'' \emph{Front. Neurosci.},
  vol.~13, p.~95, 2019.

\bibitem{deng2021optimal}
S.~Deng and S.~Gu, ``Optimal conversion of conventional artificial neural
  networks to spiking neural networks,'' in \emph{Proc. 9th Int. Conf. Learn.
  Repre.}, 2021.

\bibitem{shaban2021adaptive}
A.~Shaban, S.~S. Bezugam, and M.~Suri, ``An adaptive threshold neuron for
  recurrent spiking neural networks with nanodevice hardware implementation,''
  \emph{Nat. Commun.}, vol.~12, no. 4234, pp. 1--11, 2021.

\bibitem{perez2021neural}
N.~Perez-Nieves, V.~C. Leung, P.~L. Dragotti, and D.~F. Goodman, ``Neural
  heterogeneity promotes robust learning,'' \emph{Nat. Commun.}, vol.~12, no.
  5791, pp. 1--9, 2021.

\bibitem{lecun2015deep}
Y.~LeCun, Y.~Bengio, and G.~Hinton, ``Deep learning,'' \emph{Nature}, vol. 521,
  no. 7553, pp. 436--444, 2015.

\bibitem{jin2018hybrid}
Y.~Jin, W.~Zhang, and P.~Li, ``Hybrid macro/micro level backpropagation for
  training deep spiking neural networks,'' in \emph{Proc. Adv. Neural Inf.
  Process. Syst.}, 2018, p. 7005–7015.

\bibitem{shrestha2018slayer}
S.~B. Shrestha and G.~Orchard, ``{SLAYER}: Spike layer error reassignment in
  time,'' in \emph{Proc. Adv. Neural Inf. Process. Syst.}, 2018, p.
  1419–1428.

\bibitem{zhang2019spike}
W.~Zhang and P.~Li, ``Spike-train level backpropagation for training deep
  recurrent spiking neural networks,'' in \emph{Proc. Adv. Neural Inf. Process.
  Syst.}, 2019, pp. 7802--7813.

\bibitem{CHENG2023217}
X.~Cheng, T.~Zhang, S.~Jia, and B.~Xu, ``Meta neurons improve spiking neural
  networks for efficient spatio-temporal learning,'' \emph{Neurocomputing},
  vol. 531, pp. 217--225, 2023.

\bibitem{gerstner2014neuronal}
W.~Gerstner, W.~M. Kistler, R.~Naud, and L.~Paninski, \emph{Neuronal dynamics:
  From single neurons to networks and models of cognition}.\hskip 1em plus
  0.5em minus 0.4em\relax Cambridge University Press, 2014.

\bibitem{fang2021incorporating}
W.~Fang, Z.~Yu, Y.~Chen, T.~Masquelier, T.~Huang, and Y.~Tian, ``Incorporating
  learnable membrane time constant to enhance learning of spiking neural
  networks,'' in \emph{Proc. IEEE/CVF Conf. Comput. Vis.}, 2021, pp.
  2661--2671.

\bibitem{nakamura2000acoustical}
S.~Nakamura, K.~Hiyane, F.~Asano, T.~Nishiura, and T.~Yamada, ``Acoustical
  sound database in real environments for sound scene understanding and
  hands-free speech recognition.'' in \emph{Proc. Int. Conf. Lang. Resources
  Eval.}, 2000, pp. 965--968.

\bibitem{gu2019stca}
P.~Gu, R.~Xiao, G.~Pan, and H.~Tang, ``{STCA}: Spatio-temporal credit
  assignment with delayed feedback in deep spiking neural networks.'' in
  \emph{Proc. 28th Int. Joint Conf. Artif. Intell.}, 2019, pp. 1366--1372.

\bibitem{zhang2020temporal}
W.~Zhang and P.~Li, ``Temporal spike sequence learning via backpropagation for
  deep spiking neural networks,'' in \emph{Proc. Adv. Neural Inf. Process.
  Syst.}, 2020, pp. 12022-12033.

\bibitem{zheng2020going}
H.~Zheng, Y.~Wu, L.~Deng, Y.~Hu, and G.~Li, ``Going deeper with
  directly-trained larger spiking neural networks,'' in \emph{Proc. AAAI Conf.
  Artif. Intell.}, 2021, pp. 11062-11070.

\bibitem{sironi2018hats}
A.~Sironi, M.~Brambilla, N.~Bourdis, X.~Lagorce, and R.~Benosman, ``{HATS}:
  Histograms of averaged time surfaces for robust event-based object
  classification,'' in \emph{Proc. IEEE Conf. Comput. Vis. Pattern Recognit.},
  2018, pp. 1731--1740.

\bibitem{bi2020graph}
Y.~Bi, A.~Chadha, A.~Abbas, E.~Bourtsoulatze, and Y.~Andreopoulos,
  ``Graph-based spatio-temporal feature learning for neuromorphic vision
  sensing,'' \emph{IEEE Trans. Image Process.}, vol.~29, pp. 9084--9098, 2020.

\bibitem{ramesh2019dart}
B.~Ramesh, H.~Yang, G.~Orchard, N.~A. Le~Thi, S.~Zhang, and C.~Xiang, ``{DART}:
  distribution aware retinal transform for event-based cameras,'' \emph{IEEE
  Trans. Pattern Anal. Mach. Intell.}, vol.~42, no.~11, pp. 2767--2780, 2019.

\bibitem{liu2020unsupervised}
Q.~Liu, G.~Pan, H.~Ruan, D.~Xing, Q.~Xu, and H.~Tang, ``Unsupervised {AER}
  object recognition based on multiscale spatio-temporal features and spiking
  neurons,'' \emph{IEEE Trans. Neural Netw. Learn. Syst.}, vol.~31, no.~12, pp.
  5300--5311, 2020.

\bibitem{gjorgjieva2016computational}
J.~Gjorgjieva, G.~Drion, and E.~Marder, ``Computational implications of
  biophysical diversity and multiple timescales in neurons and synapses for
  circuit performance,'' \emph{Curr. Opin. Neurobiol.}, vol.~37, pp. 44--52,
  2016.

\bibitem{johnson2017homeostatic}
A.~P. Johnson, J.~Liu, A.~G. Millard, S.~Karim, A.~M. Tyrrell, J.~Harkin,
  J.~Timmis, L.~J. McDaid, and D.~M. Halliday, ``Homeostatic fault tolerance in
  spiking neural networks: a dynamic hardware perspective,'' \emph{{IEEE}
  Trans. Circuits Syst. {I}}, vol.~65, no.~2, pp. 687--699, 2017.

\bibitem{chen2022adaptive}
Y.~Chen, Y.~Mai, R.~Feng, and J.~Xiao, ``An adaptive threshold mechanism for
  accurate and efficient deep spiking convolutional neural networks,''
  \emph{Neurocomputing}, vol. 469, pp. 189--197, 2022.

\bibitem{cheng2020lisnn}
X.~Cheng, Y.~Hao, J.~Xu, and B.~Xu, ``{LISNN}: Improving spiking neural
  networks with lateral interactions for robust object recognition.'' in
  \emph{Proc. 29th Int. Joint Conf. Artif. Intell.}, 2020, pp. 1519--1525.

\bibitem{rokach2010ensemble}
L.~Rokach, ``Ensemble-based classifiers,'' \emph{Artif. Intell. Rev.}, vol.~33,
  pp. 1--39, 2010.

\bibitem{li2021differentiable}
Y.~Li, Y.~Guo, S.~Zhang, S.~Deng, Y.~Hai, and S.~Gu, ``Differentiable spike:
  Rethinking gradient-descent for training spiking neural networks,'' in
  \emph{Proc. Adv. Neural Inf. Process. Syst.}, vol.~34, 2021, pp. 23426-23439.

\bibitem{ghosh2009spiking}
S.~Ghosh-Dastidar and H.~Adeli, ``Spiking neural networks,'' \emph{Int. J.
  Neural Syst.}, vol.~19, no.~04, pp. 295--308, 2009.

\bibitem{tavanaei2019deep}
A.~Tavanaei, M.~Ghodrati, S.~R. Kheradpisheh, T.~Masquelier, and A.~Maida,
  ``Deep learning in spiking neural networks,'' \emph{Neural Netw.}, vol. 111,
  pp. 47--63, 2019.

\bibitem{rathi2021diet}
N.~Rathi and K.~Roy, ``{DIET-SNN}: A low-latency spiking neural network with
  direct input encoding and leakage and threshold optimization,'' \emph{IEEE
  Trans. Neural Netw. Learn. Syst.}, in press,
  {\color{blue}{doi:10.1109/TNNLS.2021.3111897}}.

\bibitem{deng2021comprehensive}
L.~Deng, Y.~Wu, Y.~Hu, L.~Liang, G.~Li, X.~Hu, Y.~Ding, P.~Li, and Y.~Xie,
  ``Comprehensive {SNN} compression using {ADMM} optimization and activity
  regularization,'' \emph{IEEE Trans. Neural Netw. Learn. Syst.}, in press,
  {\color{blue}{doi:10.1109/TNNLS.2021.3109064}}.

\bibitem{zhu2022training}
Y.~Zhu, Z.~Yu, W.~Fang, X.~Xie, T.~Huang, and T.~Masquelier, ``Training spiking
  neural networks with event-driven backpropagation,'' in \emph{Proc. Adv.
  Neural Inf. Process. Syst.}, 2022.

\bibitem{guoloss}
Y.~Guo, Y.~Chen, L.~Zhang, X.~Liu, Y.~Wang, X.~Huang, and Z.~Ma, ``{IM-Loss}:
  Information maximization loss for spiking neural networks,'' in \emph{Proc.
  Adv. Neural Inf. Process. Syst.}, 2022.

\bibitem{li2022ic}
C.~Li, Z.~Shang, L.~Shi, W.~Gao, and S.~Zhang, ``{IC-SNN}: Optimal {ANN2SNN}
  conversion at low latency,'' \emph{Mathematics}, vol.~11, no.~1, p.~58, 2022.

\bibitem{wang2022signed}
Y.~Wang, M.~Zhang, Y.~Chen, and H.~Qu, ``Signed neuron with memory: Towards
  simple, accurate and high-efficient {ANN-SNN} conversion,'' in \emph{Proc.
  31th Int. Joint Conf. Artif. Intell.}, 2022.

\bibitem{zhang2020intrinsic}
A.~Zhang, Y.~Gao, Y.~Niu, X.~Li, and Q.~Chen, ``Intrinsic plasticity for online
  unsupervised learning based on soft-reset spiking neuron model,'' \emph{IEEE
  Trans. Cogn. Dev. Syst.}, vol.~14, no.~8, pp. 1--11, 2020.

\bibitem{zhang2019fast}
A.~Zhang, H.~Zhou, X.~Li, and W.~Zhu, ``Fast and robust learning in spiking
  feed-forward neural networks based on intrinsic plasticity mechanism,''
  \emph{Neurocomputing}, vol. 365, pp. 102--112, 2019.

\bibitem{nickel2010evolutionary}
M.~Nickel, ``Evolutionary emergence of synaptic nervous systems: what can we
  learn from the non-synaptic, nerveless porifera?'' \emph{Invertebr. Biol.},
  vol. 129, no.~1, pp. 1--16, 2010.

\bibitem{rubinov2011neurobiologically}
M.~Rubinov, O.~Sporns, J.-P. Thivierge, and M.~Breakspear, ``Neurobiologically
  realistic determinants of self-organized criticality in networks of spiking
  neurons,'' \emph{PLoS Comput. Biol.}, vol.~7, no.~6, p. e1002038, 2011.

\bibitem{azouz2003adaptive}
R.~Azouz and C.~M. Gray, ``Adaptive coincidence detection and dynamic gain
  control in visual cortical neurons in vivo,'' \emph{Neuron}, vol.~37, no.~3,
  pp. 513--523, 2003.

\bibitem{marsat2010neural}
G.~Marsat and L.~Maler, ``Neural heterogeneity and efficient population codes
  for communication signals,'' \emph{J. Neurophysiol.}, vol. 104, no.~5, pp.
  2543--2555, 2010.

\bibitem{kilpatrick2013optimizing}
Z.~P. Kilpatrick, B.~Ermentrout, and B.~Doiron, ``Optimizing working memory
  with heterogeneity of recurrent cortical excitation,'' \emph{J. Neurosci.},
  vol.~33, no. 48, pp. 18999--19011, 2013.

\bibitem{spyrou2021neuron}
T.~Spyrou, S.~A. El-Sayed, E.~Afacan, L.~A. Camu{\~n}as-Mesa,
  B.~Linares-Barranco, and H.-G. Stratigopoulos, ``Neuron fault tolerance in
  spiking neural networks,'' in \emph{Proc. of DATE}, 2021, pp. 743--748.

\bibitem{pellegrini2021low}
T.~Pellegrini, R.~Zimmer, and T.~Masquelier, ``Low-activity supervised
  convolutional spiking neural networks applied to speech commands
  recognition,'' in \emph{Proc. IEEE Spok. Lang. Technol. Workshop}, 2021, pp.
  97--103.

\bibitem{meng2022training}
Q.~Meng, M.~Xiao, S.~Yan, Y.~Wang, Z.~Lin, and Z.-Q. Luo, ``Training
  high-performance low-latency spiking neural networks by differentiation on
  spike representation,'' in \emph{Proc. IEEE/CVF Conf. Comput. Vis. Pattern
  Recognit.}, 2022, pp. 12444-12453.

\bibitem{jiang2021few}
R.~Jiang, J.~Zhang, R.~Yan, and H.~Tang, ``Few-shot learning in spiking neural
  networks by multi-timescale optimization,'' \emph{Neural Comput.}, vol.~33,
  no.~9, pp. 2439--2472, 2021.

\end{thebibliography}

	\bibliographystyle{IEEEtran}

\end{document}